\documentclass[10pt,twocolumn,letterpaper]{article}

\usepackage{3dv}
\usepackage{times}
\usepackage{epsfig}
\usepackage{graphicx}
\usepackage{amsmath}
\usepackage{amssymb}
\usepackage{bm}

\usepackage{booktabs}
\usepackage{mwe}
\usepackage{colortbl}
\usepackage{multirow}
\usepackage{color}
\usepackage[dvipsnames]{xcolor}
\usepackage{arydshln}
\usepackage{stmaryrd}
\usepackage{gensymb}

\usepackage{enumitem}
\setlist{nosep}

\newcommand{\PreserveBackslash}[1]{\let\temp=\\#1\let\\=\temp}
\newcolumntype{C}[1]{>{\PreserveBackslash\centering}p{#1}}
\newcolumntype{R}[1]{>{\PreserveBackslash\raggedleft}p{#1}}
\newcolumntype{L}[1]{>{\PreserveBackslash\raggedright}p{#1}}

\newcommand\was[1]{}

\definecolor{mygray}{rgb}{0.5, 0.5, 0.5}
\definecolor{DarkBlue}{rgb}{0.0, 0.5, 0.8}

\iftrue
\let\renaud\relax
\let\rmf\relax
\let\update\relax
\let\yang\relax
\else
\newcommand\renaud{\textcolor{orange}}
\newcommand\yang{\textcolor{DarkBlue}}
\newcommand\rmf{\textcolor{orange}}
\newcommand\update{\textcolor{DarkBlue}}
\fi

\definecolor{HL}{rgb}{0.95, 1.0, 0.95}

\newcommand{\loss}{\mathcal{L}}
\newcommand{\feat}{f}
\newcommand{\img}{I}

\newcommand{\bin}{B}
\newcommand{\azi}{\alpha}
\newcommand{\ele}{\beta}
\newcommand{\rot}{\gamma}
\newcommand{\euler}{\theta}
\newcommand{\pose}{\mathbf{R}}
\newcommand{\encoder}{Enc}
\newcommand{\predictor}{Pred}

\newcommand{\dist}{\mathrm{d}}
\newcommand{\AccThirty}{\text{Acc30}}
\newcommand{\MedErr}{\text{MedErr}}
\newcommand{\ang}{{\text{ang}}}
\newcommand{\infoNCE}{{\text{infoNCE}}}
\newcommand{\poseNCE}{{\text{poseNCE}}}

\makeatletter
\renewcommand\paragraph{\@startsection{paragraph}{4}{\z@}%
                                    {1ex \@plus.8ex \@minus.2ex}
                                    {-1em}%
                                    {\normalfont\normalsize\bfseries}}
\makeatother

\usepackage[pagebackref=true,breaklinks=true,letterpaper=true,colorlinks,bookmarks=false]{hyperref}

\threedvfinalcopy 

\def\threedvPaperID{41} 

\ifthreedvfinal\fi

\begin{document}

\title{
PoseContrast: Class-Agnostic Object Viewpoint Estimation in the Wild \\ with Pose-Aware Contrastive Learning
}

\author{Yang Xiao$^1$ \qquad Yuming Du$^1$ \qquad Renaud Marlet$^{1,2}$ \\
\\
\hspace{-3mm}$^1$LIGM, Ecole des Ponts, Univ Gustave Eiffel, CNRS, Marne-la-Vallée, France \hspace{1mm} $^2$Valeo.ai, Paris, France
}

\maketitle

\thispagestyle{empty} 

\begin{abstract}

Motivated by the need for estimating the 3D pose of arbitrary objects, we consider the challenging problem of class-agnostic object viewpoint estimation from images only, without CAD model knowledge. The idea is to leverage features learned on seen classes to estimate the pose for classes that are unseen, yet that share similar geometries and canonical frames with seen classes. We train a direct pose estimator in a class-agnostic way by sharing weights across all object classes, and we introduce a contrastive learning method that has three main ingredients: (i)~the use of pre-trained, self-supervised, contrast-based features; (ii)~pose-aware data augmentations; (iii)~a pose-aware contrastive loss. We experimented on Pascal3D+, ObjectNet3D and Pix3D in a cross-dataset fashion, with both seen and unseen classes. We report state-of-the-art results, including against methods that additionally use CAD models as input.
Code is available at \url{https://github.com/YoungXIAO13/PoseContrast}.

\end{abstract}

\section{Introduction}
\label{sec:Intro}

\begin{figure}[t]
    \centering
    \vspace*{-2mm}
    \includegraphics[width=1\columnwidth]{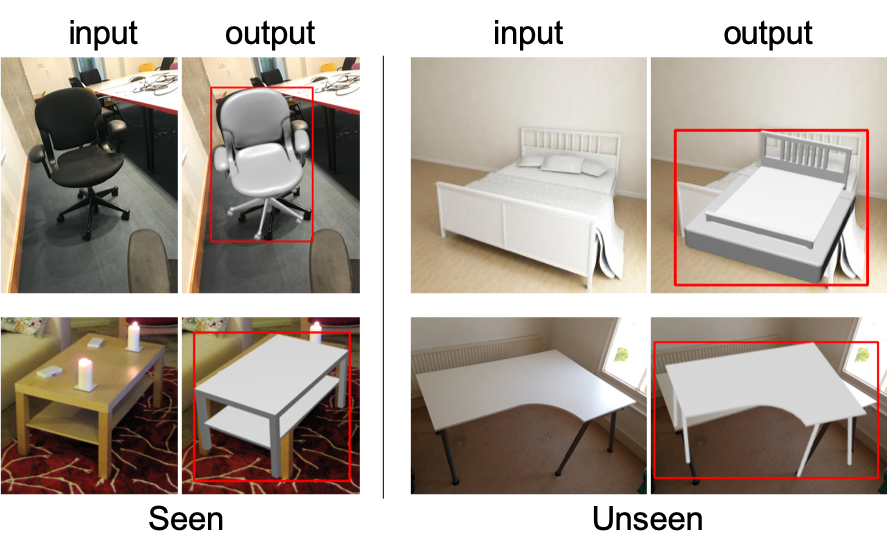}
    \vspace*{-6mm}
    \caption{{\bf Task Illustration.} 
    Given an RGB image picturing an object, we aim to estimate its 3D pose (viewpoint) without knowing its class or shape.
    It is made possible by training a model for all objects in a class-agnostic way and applying it to objects of unseen classes having similar geometries as training objects \renaud{and} similar canonical frames, e.g., an unseen desk being similar to a seen table.
    (Red boxes are detections of a class-agnostic Mask R-CNN and the 3D models here are only used to visualize the pose.)}
    \label{fig:teaserTask}
    \vspace*{-2.2mm}
\end{figure}

Object 3D pose (viewpoint) estimation aims at predicting the 3D rotation of objects in images with respect to the camera. 
Deep learning, as well as datasets containing a variety of pictured objects annotated with 3D pose, have led to great advances in this task \cite{ViewpointsKeypoints2015,Su2015RenderFC,Mousavian20163DBB,Wang_2019_NOCS,Liao2019SphericalRL}.

However, they mainly focus on class-specific estimation for few categories, and they \renaud{mostly evaluate} on ground-truth bounding boxes. It is an issue when encountering objects of unseen classes \rmf{or with out-of distribution appearance}, for which no training data was available and no bounding box \renaud{is} given, \renaud{which is a likely circumstance for robots in uncontrolled \rmf{environments, e.g., outdoors vs in factories}.}

\textbf{Our goal}
\renaud{is to address this issue. Given training data for some known classes (images with bounding boxes of multiple objects, class labels and 3D pose annotations), we want to detect and estimate the 3D pose of objects of unknown classes, given only an RGB image as input (Fig.\,\ref{fig:teaserTask}), vs also using CAD models of objects as some methods do \cite{Xiao2019PoseFS, Pitteri2020ACCV}.}

\renaud{\textbf{This new task} relies on two assumptions. First, it applies to unseen classes that share similarities with seen classes. For example, one may expect to orient an unseen bed when trained on seen chairs and sofas, but not a wrench.}

\begin{figure}[t]
    \centering
    \includegraphics[width=0.9\columnwidth]{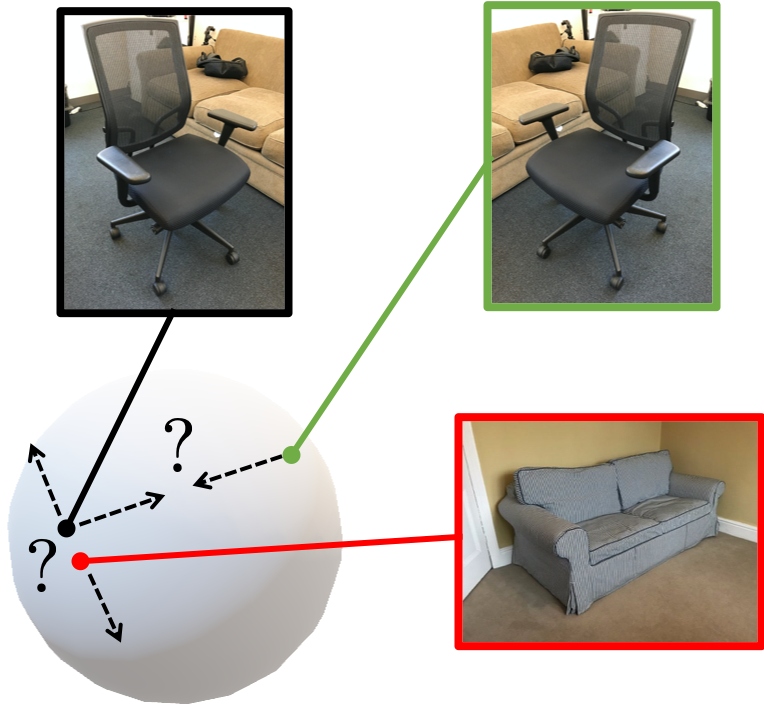}
    \caption{\textbf{Pose-Aware Contrastive Loss.} In usual self-supervised contrastive training, the network learns to pull together in feature space the {\bf query} (e.g., chair) and a \textcolor{LimeGreen}{\bf positive} variant (e.g., flipped image), while pushing apart the \textbf{query} from \textcolor{Red}{\bf negatives} (different objects, e.g., sofa), ignoring pose information.
    Instead, we exclude flipped \textcolor{LimeGreen}{\bf positives}, whose pose actually differ from the \textbf{query}, and do not push apart
    \textcolor{Red}{\bf negatives} with similar poses (e.g., sofa).
    }
    \vspace*{-1mm}
    \label{fig:teaser}
\end{figure}

\renaud{The other assumption is that similar classes have a consistent canonical pose, i.e., have aligned similarities (Figs.\,\ref{fig:teaser} and \ref{fig:PoseContrast}). It is somehow a weak assumption, satisfied by all datasets we know of, probably because many objects are used consistently w.r.t.\ verticality, and feature a notion of left- and right-hand sides, or at least a main vertical symmetry plane, which is enough to define a ``natural'' canonical frame, possibly up to symmetry. Besides, if similar classes in a training set have inconsistent canonical poses, they can be normalized by a systematic rotation; no 3D shape is needed for that. In this first work, we only consider the general case, disregarding the different forms of symmetry.}

\renaud{\textbf{Overview.}} \renaud{To detect arbitrary objects and estimate their pose, although not in training data, we use a class-agnostic approach for both object detection and pose estimation.}

Approaches like \cite{Grabner20183DPE, zhou2018starmap, Pitteri2019CorNetG3} have already demonstrated the effectiveness of this setting. They detect 2D keypoints regardless of the class of the object, estimate 2D-3D keypoint correspondences, and use a PnP algorithm~\cite{Lepetit2008EPnPAA} to compute the pose. But besides being indirect, these methods need a suitable design of class-agnostic keypoints on various object geometries.
\yang{In contrast, our approach estimates the 3D pose \renaud{directly} from the image embeddings, without any intermediate representation.}

Others assume a 3D model of the object is given at test time (sometimes also at training time)~\cite{Xiao2019PoseFS, park2020latent, Pitteri2020ACCV, Dani_2021_WACV}, either provided by a human or retrieved automatically by an algorithm, which is hard due to the image-shape domain gap and to the number of classes to discriminate~\cite{Su2015MultiviewCN,Massa2015DeepE2,xiang2016objectnet3d}, and it is limited by the database of possible 3D models to handle. \yang{In comparison, our method relies only on RGB images \renaud{both at train and test time}, without any \renaud{CAD model as input}.}

To that end, we train a class-agnostic pose estimator by sharing weights across all object classes. And we propose a contrastive-learning approach to learn geometry-aware image embeddings that are optimized for pose estimation.

Recent contrastive-learning approaches create discriminative image features by learning to distinguish pairs of \emph{identical objects} with different appearances \renaud{thanks} to a synthetic transformation (positives), from pairs of \emph{different objects} (negatives). 
Inspired by image-level discrimination~\cite{He2020moco,chen2020simCLR,xiao2021what}, we adapt the common contrastive loss InfoNCE~\cite{Oord2018infoNCE} \renaud{so that it discriminates poses rather than categories: we propose PoseNCE,} a pose-aware contrastive loss that pushes away in latent space the image features of objects having different poses, ignoring the class of these objects as we aim at class-agnostic pose estimation \renaud{(see Fig.\,\ref{fig:teaser})}.
Besides, \renaud{departing from the binary separation between positives and negatives in classical InfoNCE}, \renaud{PoseNCE} takes into account the level of pose difference \renaud{between two objects} as a weighting term to reduce or stress the negativeness of a pair, regardless of the class (see Fig.\,\ref{fig:PoseContrast}).

Concretely, we use both an angle loss and a contrastive loss. We also curate the contrastive learning transformations to distinguish pose-variant data augmentations, e.g., horizontal flip, and pose-invariant data augmentations, e.g., color jittering. The former is used to actually augment the dataset, while the latter is used to create similar variants to construct positive and negative pairs. 
And rather than training from scratch on available data\-sets, that are relatively small, we initialize our network with a contrastive model trained on a large dataset in a self-supervised way.

\renaud{Last}, we propose a class-agnostic approach for \renaud{both} object detection \emph{and} pose estimation. For this, we train a Mask R-CNN in a class-agnostic way for generic object detection\renaud{, and pipeline} it with our pose estimator, thus addressing the \renaud{coupled} problem of generic object detection and pose estimation for unseen objects. It is a more realistic setting w.r.t.\ existing class-agnostic pose estimation methods, that only evaluate in the ideal case of ground-truth bounding boxes.

\renaud{\textbf{Our main contributions} are as follows:}
\begin{itemize}[itemsep=1pt,topsep=0pt]

    \item \renaud{We define a new task suited for uncontrolled settings: class-agnostic object 3D pose estimation, possibly coupled and preceded by class-agnostic detection.}

    \item \renaud{We propose a contrastive-learning approach for class-agnostic pose estimation, which includes a pose-aware contrastive loss and pose-aware data augmentations.}

    \item \renaud{We report state-of-the-art results on 3 datasets, including against methods that also require shape knowledge. (And code will be made public upon publication.)}
\end{itemize}

\section{Related Work}
\label{sec:Rela}

\paragraph{Class-Specific Object Pose Estimation.}
While instance-level 3D object pose estimation has long been studied in both robotic and vision communities~\cite{Hinterstoier2012ModelBT,Brachmann2014Learning6O,Kehl2017SSD6DMR,Rad2017BB8AS,Tekin2017RealTimeSS,Rad2018FeatureMF,xiang2018posecnn,Sundermeyer2018Implicit3O,Oberweger2018MakingDH,labbe2020}, class-level pose estimation has developed more recently thanks to learning-based methods~\cite{Su2015RenderFC,ViewpointsKeypoints2015,Tulsiani2015PoseIF,Mousavian20163DBB,Kundu20183DRCNNI3,wang20183d,Grabner20183DPE,Grabner2019GP2CGP,Wang_2019_NOCS,zhou2018starmap,Tseng2019FewShotVE}. These methods can be roughly divided into two categories: pose estimation methods that regress 3D orientations directly 
\cite{ViewpointsKeypoints2015,Su2015RenderFC,Mousavian20163DBB,wang20183d,Xiao2019PoseFS}, and keypoint-based methods that predict 2D locations of 3D keypoints~\cite{Grabner20183DPE,Grabner2019GP2CGP,Wang_2019_NOCS,zhou2018starmap,Tseng2019FewShotVE}.

Still, annotating 3D pose for objects in the wild is a tedious process of searching best-matching CAD models and aligning them to images~\cite{xiang2014pascal3d,xiang2016objectnet3d}. It does not scale to large numbers of objects and classes. While good performance is achieved on supervised classes, generalizing beyond training data remains an important, yet under-explored problem.

\paragraph{Class-Agnostic Object Pose Estimation.}

To circumvent the problem of limited labeled object classes, a few class-agnostic pose estimation methods have recently been proposed~\cite{Ge2020pose, Grabner20183DPE, zhou2018starmap, Xiao2019PoseFS, Dani_2021_WACV, Pitteri2019CorNetG3, Pitteri2020ACCV}. In contrast to class-specific methods that build an independent prediction branch for each object class, agnostic methods estimate the object pose without knowing its class {\it a priori}, which is enabled by sharing model weights across all object classes during training.

\renaud{\cite{Ge2020pose} trains on multiple views of the same object instance on a turntable.}
\cite{Grabner20183DPE,Pitteri2019CorNetG3} use the 3D bounding box corners as generic keypoints for class-agnostic object pose estimation. However, \cite{Grabner20183DPE} only reports performance on seen classes and \cite{Pitteri2019CorNetG3} focuses on cubic objects with simple geometric shape. Instead of using a fixed set of keypoints for all objects, \cite{zhou2018starmap} propose a class-agnostic keypoint-based approach combining a 2D keypoint heatmap and 3D keypoint locations in the object canonical frame. 
These methods are robust on textured objects but fail with heavy occlusions and tiny or textureless objects. In contrast, our method ignores keypoints, directly infers a pose and is less sensitive to texturelessness.

Rather than relying only on RGB images, another group of class-agnostic pose estimation methods~\cite{Xiao2019PoseFS, Dani_2021_WACV, Pitteri2020ACCV} use 3D models, in particular at test time to adapt to objects unseen at training time. \cite{Xiao2019PoseFS} aggregates 3D shape and 2D image information for arbitrary objects, representing 3D shapes as multi-view renderings or point clouds. \cite{Dani_2021_WACV} proposes a lighter version of~\cite{Xiao2019PoseFS} encoding the 3D models into graphs using node embeddings~\cite{Grover2016node2vecSF}. 
\renaud{\cite{Pitteri2020ACCV} matches local images embeddings with local 3D embeddings, then use RANSAC and PnP algorithms to recover an object from a database of CAD models, and a pose.}
In contrast, we need no 3D shape, neither at training nor at testing time.


\was{
There are also methods working on self-supervised pose estimation by learning directly from unlabeled images~\cite{deng2020self,Wang2020Self6DSM,mustikovelaCVPR20selfView} but they focus on one or several seen classes.
}

\paragraph{\renaud{Pose Loss.}}

\renaud{3D pose dissimilarity has been measured indirectly, e.g., with a distance on reprojected features such as keypoints (see above), or directly on pose parameters. In the latter case, the chosen representation and penalty may yield more or less artefacts due to, e.g., discontinuities in the parameterization (Euler angles, quaternions \cite{Zhou2019Continuity}), gimbal lock \cite{Grassia1998practical}, anti-podal symmetry (quaternions), non-uniform parameter distributions, classification discretization \cite{ViewpointsKeypoints2015, Su2015RenderFC, Elhoseiny2016ACA}, single-mode analysis as with regression \cite{Osadchy2007Synergistic, Penedones2011Improving, Massa2016CraftingAM}, or parameter-space biases when penalizing with the L2-norm of the difference of pose parameters, including with  the exponential twist representation \cite{Zhu2017rethinking}. We use a combination of classification and regression \cite{Mousavian20163DBB, Guler2017DenseReg, MahendranBMVC2018, UnifiedMVMC2018} of Euler angles similar to \cite{Xiao2019PoseFS} (offset regression from bin center), which better separates modes in case of pose ambiguities, but we penalize a geodesic distance on the unit sphere rather than the Euclidean distance of parameters, which does not have dimensional biases.}

\paragraph{Contrastive Learning.}

Instead of designing pretext tasks for unsupervised learning~\cite{Doersch2015UnsupervisedVR,Noroozi2016UnsupervisedLO,Zhang2016ColorfulIC,gidaris2018unsupervised}, 
powerful image features can be learned by contrasting positive and negative pairs~\cite{Wu2018UnsupervisedFL,Oord2018infoNCE,tian2019contrastive,Misra2020PIRL,caron2020swav,He2020moco,chen2020mocov2,chen2020simCLR,chen2020simCLRv2,khosla2020SupervisedCL}. Among the various forms of the contrastive loss function~\cite{Hadsell2006DimensionalityRB,Wang2015UnsupervisedLO,hjelm2018learning,Wu2018UnsupervisedFL,Oord2018infoNCE}, InfoNCE~\cite{Oord2018infoNCE} has become a standard pick in many methods.

While most contrastive-learning approaches work in the unsupervised setting, \cite{khosla2020SupervisedCL} operates with full supervision. Considering the class label of training examples, features belonging to the same class are pulled together while features from different classes are pushed apart.

Similar to~\cite{khosla2020SupervisedCL}, we also propose a contrastive loss that works in the fully-supervised setting. However, instead of focusing on semantic label information, we design it for our geometric task, taking into account the pose distance between different examples. Moreover, we also curate data augmentations as advocated in~\cite{xiao2021what}, leaving out those that would be harmful for our pose estimation task.

\renaud{Besides requiring 3D shapes at training time and operating on RGB-D data, \cite{Balntas2017PoseGR} is not pose-aware: in the InfoNCE spirit, it creates positive pairs from the same known shape model and negative pairs from known different shapes, ignoring pose. Besides, it favors features whose L2-distance is \emph{equal to} their pose L2-distance, which is a heavy burden for feature learning, especially for objects with large shape variations. In comparison, we simply contrast features w.r.t.\ pose dissimilarity. \cite{Wohlhart2015LearningDF}, which operates in a class-specific way and also requires known 3D shapes or at least multiple views or renderings of each object, uses a triplet loss whose formulation can be related to our more general PoseNCE loss, but it does not take into account the level of pose dissimilarity nor pose-aware data augmentation.}


\paragraph{\renaud{Coupled Detection and Pose Estimation.\!\!}}

\renaud{\!\!Very few works consider the realistic scenario of detecting unknown objects in images \emph{and} inferring their pose. \cite{Pitteri2020ACCV} trains a class-agnostic Mask R-CNN and pipelines it with a pose estimator, as we do, but it applies to industrial objects and requires knowing the 3D shapes, including for novel instances. \cite{Ge2020pose}, which trains with objects on a turntable, does not do any detection but somehow also applies to ImageNet, i.e., with well-centered, single-object images. None of these methods is thus applicable to objects in the wild. And although \cite{Grabner20183DPE} predicts a 3D box size (not location) for PnP reprojection, it operates on ground-truth 2D bounding boxes.
We can only compare in the class-specific detection and pose estimation setting \cite{wang20183d, Grabner2019GP2CGP} and, in the class-agnostic setting, against methods also requiring an input 3D shape \cite{Xiao2019PoseFS}.}

\section{Method}
\label{sec:Method}


\begin{figure}[t]
    \centering
    \includegraphics[width=0.95\columnwidth]{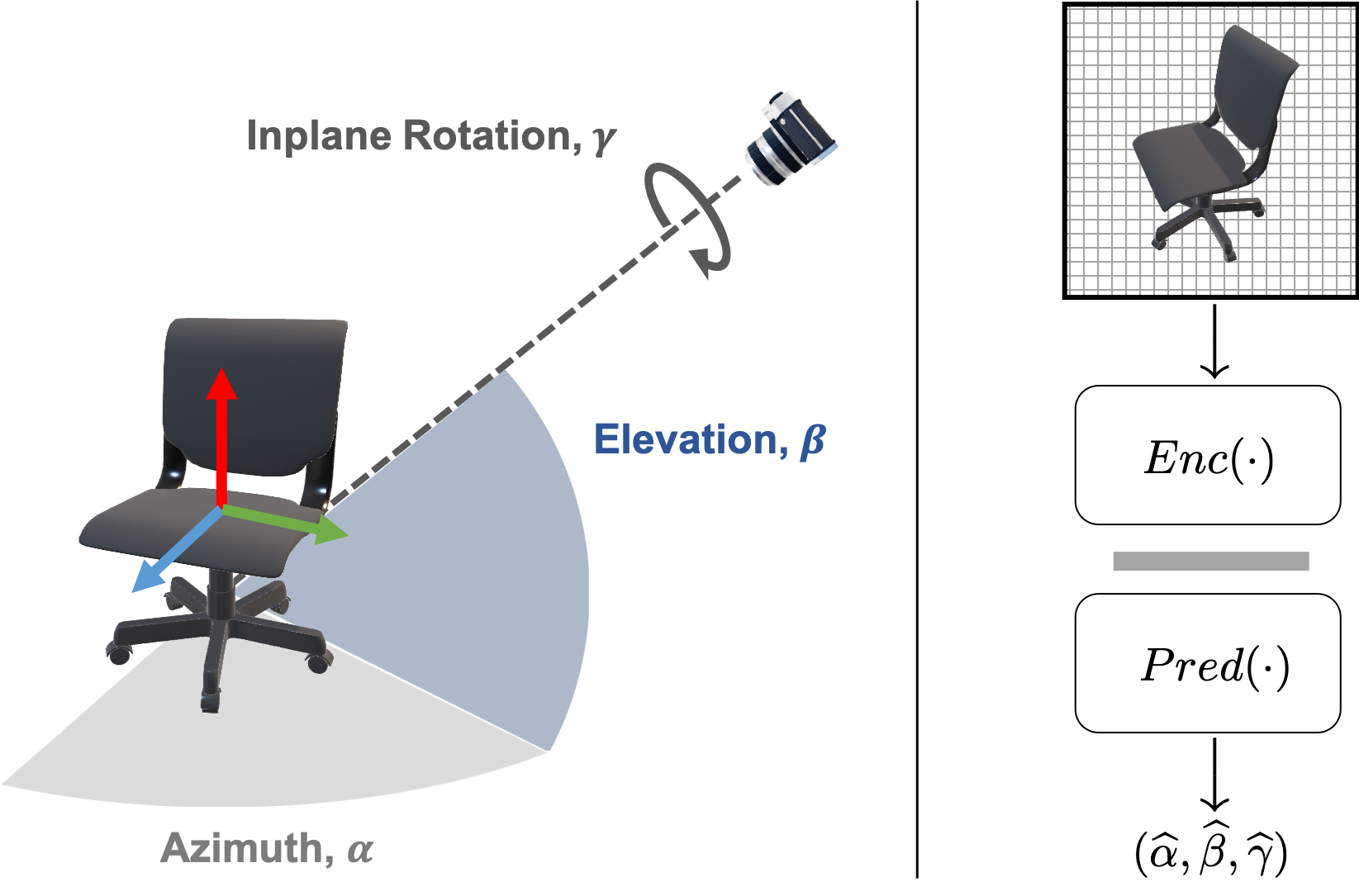}
    \caption{\textbf{Pose Parameters} (left). \textbf{Network Architecture} (right): from an image crop, the encoder $\encoder$ produces an embedding, which is given to the predictor $\predictor$ to produce pose angles.}
    \label{fig:network}
    \vspace*{-1mm}
\end{figure}

Given an RGB image $\img$ containing an object at a given (known or detected) image location, we aim to estimate the 3D pose $\pose$ of the object with no prior knowledge of its class or shape. To that end, we crop the image region containing the object, encode it to produce class-agnostic features, from which the object 3D pose is directly predicted (Fig.\,\ref{fig:network}).

\paragraph{3D Pose Parameterization.}

To predict the 3D rotation matrix $\pose$ of the pictured object, we decompose it into three Euler angles as in \cite{Su2015RenderFC,Xiao2019PoseFS}: azimuth $\azi$, elevation $\ele$, and inplane rotation~$\rot$, with $\azi, \rot \in [-\pi, \pi)$ and  $\ele \in [-\pi/2, \pi/2]$.

Recent work on pose estimation shows a higher performance
with a \rmf{more continuous} formulation (cf.~\cite{Zhou2019Continuity}) mixing angular bin classification and within-bin offset regression \cite{Xiao2019PoseFS, Xiao2020FewShotOD}.
Concretely, we split each Euler angle $\euler \,{\in}\, \{\azi, \ele, \rot\}$ uniformly into discrete bins $i$ of size $\bin$ ($= \pi / 12$ in our experiments).
The network outputs bin classification scores $p_{\euler, i} \,{\in}\, [0, 1]$ and offsets $\delta_{\euler, i} \,{\in}\, [0, 1]$ within the bin. 

\paragraph{Angle Loss.}

We use a cross-entropy loss for angle bin classification and a smooth-L1 loss for bin offset regression:
\begin{equation} \label{eq:PoseLoss}
    \loss_\ang = \sum_{\theta \in \{ \azi, \ele, \rot \}} \loss_\mathrm{cls} (\mathrm{bin}_\theta, p_{\euler}) + \lambda \, \loss_\mathrm{reg} (\mathrm{offset}_\theta, \delta_{\euler})
\end{equation}
where $\mathrm{bin}_\theta$ is the ground-truth bin and $\mathrm{offset}_\theta$ is the offset for angle $\theta$. The relative weight $\lambda$ is set to 1 in our experiments. The final prediction for angle $\euler$ is obtained as:
\begin{equation} \label{eq:EulerAngles}
    \widehat{\euler} = (j + \delta_{\euler, j}) \bin
    \quad \mathrm{with} \quad 
    j = \arg \max_i p_{\euler, i}
\end{equation}
where $i\,{\in}\, [-12..11]$ for $\azi, \rot$, and $i \,{\in}\, [-6..5]$ for $\ele$. The angle loss is complemented by a contrastive loss (cf.\ Sect.\,\ref{sec:ContrastiveLoss})\rlap.

\paragraph{Network Architecture.}

The architecture of our network is depicted in Figure~\ref{fig:network} (right). It consists of two modules: an image encoder $\encoder(\cdot)$ and a pose predictor $\predictor(\cdot)$.

For feature extraction, we use a standard CNN, namely ResNet-50. We crop the input image to the targeted object and pass it through the encoder network until the max-pooling layer. It provides a 2048-dimension feature vector.

We then pass the image embedding through the pose predictor, which is a multi-layer perceptron (MLP) with 3 hidden layers of size 800-400-200, each followed by batch normalization and ReLU activation. Contrary to class-specific methods~\cite{Su2015RenderFC,ViewpointsKeypoints2015,Liao2019SphericalRL,Mousavian20163DBB} that use one prediction branch per class, we use a single prediction branch for all objects.


\label{sec:ContrastiveLoss}



\begin{figure}[t]
    \centering
    \vspace*{-2mm}
    \includegraphics[width=0.9\columnwidth]{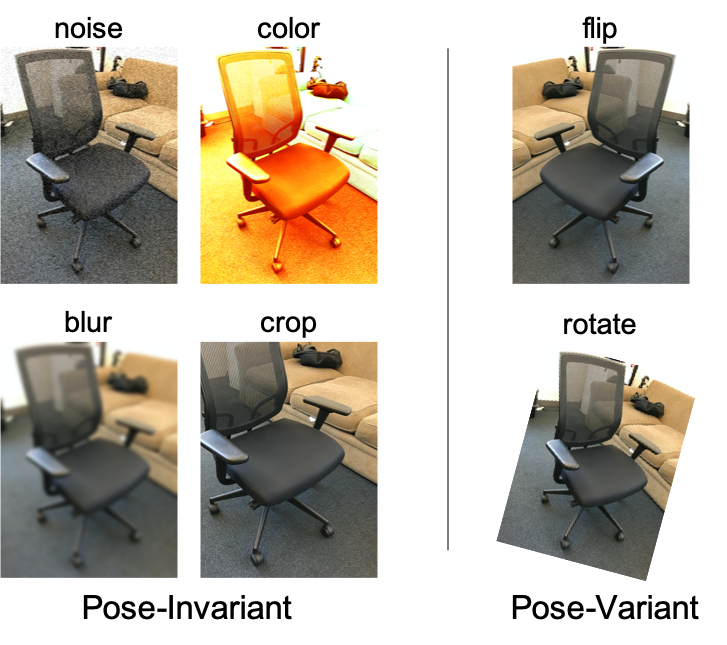}
    \vspace*{-1mm}
    \caption{\textbf{Pose-Aware Data Augmentations:} 
    while \emph{pose-invariant} data augmentations do not alter the pose, \emph{pose-variant} augmentations modify it and cannot be used as positives.}
    \label{fig:DataAug}
    \vspace*{-2mm}
\end{figure}

\paragraph{Contrastive Features.}


Datasets of images with pose annotations are scarce and small. (One of the reasons is probably that pose is much harder to annotate than class, especially for images in the wild.) It makes it difficult to learn a high-quality pose estimator. Rather than learning a network from scratch, as most other methods do, or from an initial ImageNet classifier, whose bias is not particularly suited for pose estimation, we initialize our predictor using a pre-trained contrast-based network \cite{chen2020mocov2}. We show that it plays a significant role in our high performance (cf.\ Sect.\,\ref{sec:expAblation}).

\paragraph{Self-Supervised Contrastive Loss.}

In self-supervised con\-trastive learning~\cite{chen2020simCLR,He2020moco}, the contrastive loss serves as an unsupervised objective for training an image encoder that maximizes agreement between different transformations of the same sample, while minimizing the agreement with other samples. Concretely, we consider a batch $(\img_k)_{k\in[1..N]}$ of training samples, transformed into $(\tilde\img_k)_{k\in[1..N]}$ by data augmentation, and encoded as $\feat_k \,{=}\, \encoder(\tilde\img_k)$. 
For any index $k^+\,{\in}\,[1..N]$, we consider an alternative augmentation $\tilde\img_q$ of $\img_{k^+}$ (the query), with embedding $\feat_q \,{=}\, \encoder(\tilde\img_q)$, and we separate the \emph{positive pair} $(q,k^+)$ from the \emph{negative pairs} $(q,k^-)_{k^-\in[1..N] \setminus\{k^+\}}$ with the following InfoNCE loss:
\begin{equation} \label{eq:infoNCE}
    \loss_\infoNCE = - \log \frac{ \exp(\feat_q {\cdot} \feat_{k^+} / \tau) }{ \sum_{k \in [1..N]} \exp(\feat_q {\cdot} \feat_{k} / \tau) }
\end{equation}
where $\tau$ is a temperature parameter~\cite{chen2020simCLR} ($0.5$ in experiments). 


\begin{figure}
    \centering
    \includegraphics[width=0.95\columnwidth]{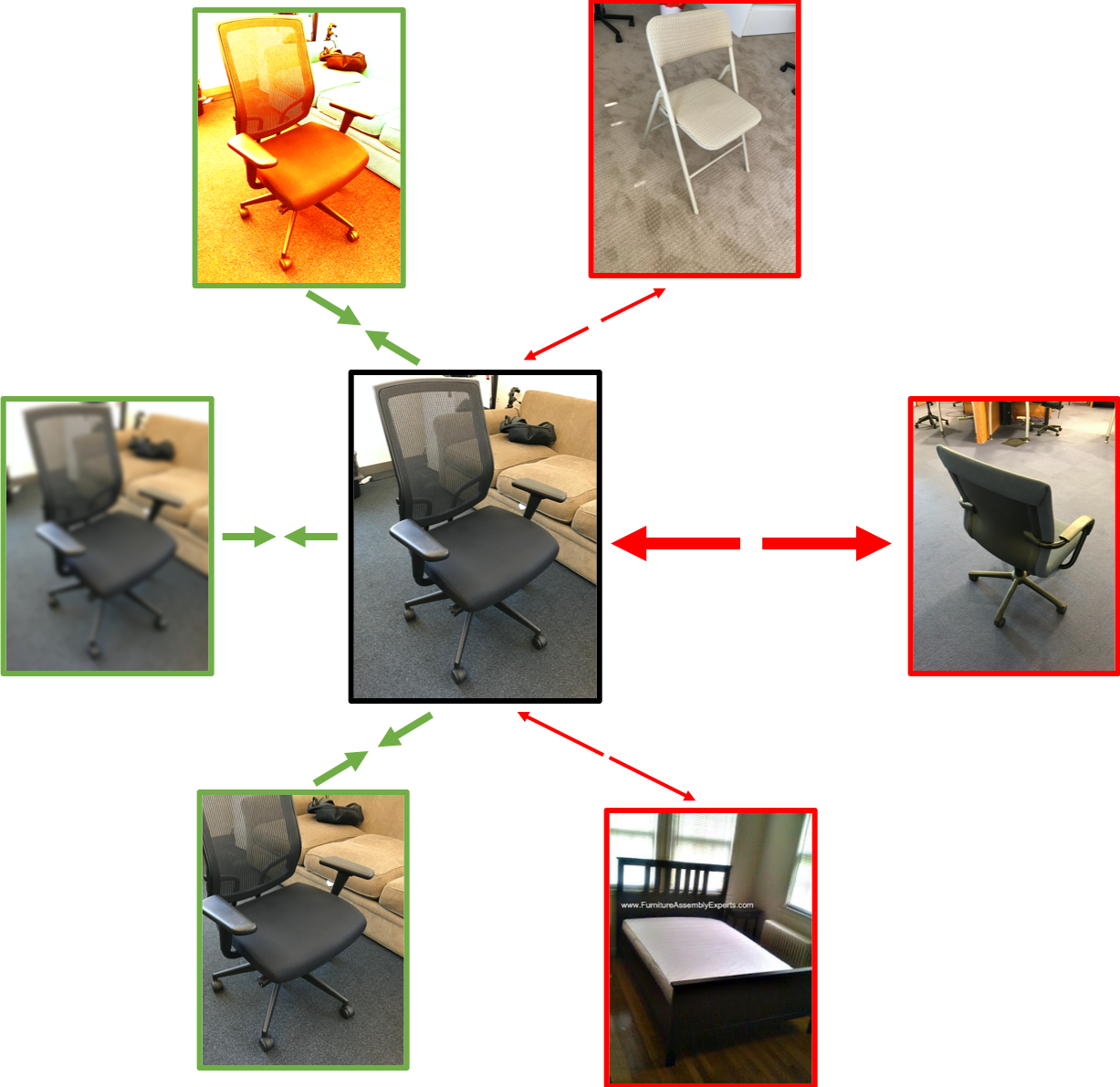}
    \caption{{\bf Pose-Aware Contrast.} Instead of treating all \textcolor{red}{negatives} (right images) equally, as for \textcolor{LimeGreen}{positives} (left images), we give more weight to \textcolor{red}{\bf negatives} with a large rotation (rightmost) and less to those with a small rotation, \emph{regardless of their semantic class}.}
    \label{fig:PoseContrast}
    \vspace*{-2mm}
\end{figure}

\paragraph{Pose-Aware Data Augmentations.}

As illustrated in Figure~\ref{fig:DataAug}, we divide data augmentations into two categories. \emph{Pose-invariant} augmentations transform the image without changing the 3D pose of objects: color jittering, blur, crop, etc. On the contrary, \emph{pose-variant} augmentations change the 3D pose at the same time: rotation and horizontal flip. More precisely, an image rotation of angle $\phi$ corresponds to an in-plane rotation of angle $\rot+\phi$ for the object, and a horizontal image flip corresponds to a change of sign of azimuth~$\azi$ and in-plane rotation~$\rot$. (We assume mirror-imaged objects are realistic objects with identical canonical frame.) 
In our experiments, $\phi$ varies in $[-15\degree,15\degree]$ \update{since 95\% of the images in Pascal3D+~\cite{xiang2014pascal3d} fall in this range}.


Unlike self-supervised learning methods that make use of all data augmentation techniques at the same time, we distinguish two augmentation times. At \emph{batch creation time}, we only apply pose-variant data augmentations, i.e., a small rotation or a horizontal flip, and update the object pose information accordingly. At \emph{contrast time}, i.e., when creating \emph{positive} and \emph{negative} pairs, we only apply pose-invariant data augmentations. The latter is motivated by \cite{xiao2021what}: a blind use of any data augmentation could be harmful.


\paragraph{Pose-Aware Contrastive Loss.}

The contrastive loss in Eq.\,\eqref{eq:infoNCE} is designed for unsupervised learning with no annotation involved. While efficient for learning instance-discriminative image embeddings, it is not particularly suited for contrasting geometric cues towards pose estimation: 
the query embedding is contrasted away from the embeddings of negative samples even if their pose is identical or similar to \renaud{the pose of} the query object.
While the case of different views of identical or similar instances can be disregarded in usual contrastive learning because of its practical rarity, similar and even identical poses are common in a single batch.  What we want, 
instead of contrasting the object semantics, is to learn pose-variant image features.

We thus introduce a new pose-aware contrastive loss, illustrated in Figure~\ref{fig:PoseContrast}. It takes into account the level of sample negativeness: the larger the pose difference, the higher the weight in the loss. \renaud{(There is thus no need to define a notion of \emph{pose similarity}.)}
Concretely, for each pair $(q,k)$, we compute a normalized distance $\dist(\pose_q, \pose_k) \,{\in}\, [0,1]$ between the associated 3D pose rotations $\pose_q, \pose_k$ and we use it as a weight in our \emph{pose-aware contrastive loss PoseNCE}:
\begin{equation}
\label{eq:poseNCE}
    \loss_\poseNCE = - \log \frac{ \exp(\feat_q {\cdot} \feat_{k^+} / \tau) }{ \sum_{k \in [1..N]} \dist(\pose_q, \pose_k)  \exp(\feat_q {\cdot} \feat_k / \tau) }
\end{equation}
with $\dist(\pose_q, \pose_{k^{+}}\!) \,{=}\, 0$ as $\pose_q \,{=}\,\pose_{k^+}$. Our distance is defined as the normalized angle difference between the rotations, which is akin to a geodesic distance on the unit sphere:
\begin{equation}
\label{eq:poseNCEdist}
\begin{aligned}
    \dist(\pose_q, \pose_k) &= \Delta(\pose_q, \pose_k) / \pi \quad \text{with}\\
    \Delta(\pose_q, \pose_k) &= \arccos \big( \frac{ \mathrm{tr}(\pose_q^{\!\mathrm{T}} \,\pose_k) - 1 }{2} \big)
    \end{aligned}
\end{equation}
Following \cite{zhang2021rethinking}, PoseNCE can be seen as softer or smoother version of InfoNCE~\cite{Oord2018infoNCE}, which itself is softer than hard pairwise or triplet losses. It can also be seen as a soft treatment of easy and hard negatives \cite{wang2020xbm}.

Last, the total loss adds the angle and contrastive losses: 
\begin{equation}
\label{eq:globalloss}
    \loss = \loss_\ang + \kappa \,\loss_\poseNCE 
\end{equation}
The relative weight $\kappa$ is set to 1 in our experiments.


\section{Experiments}
\label{sec:exp}

In this section, we introduce our experimental setup\renaud{, analyze results on three commonly-used datasets, provide an ablation study and discuss the limitations.}
%
\renaud{The supplementary material include details on datasets, splits, training, implementation, and more classwise results. It also provides a study on the temperature parameter, and more visual results.}





\begin{table*}[!t]
\addtolength{\tabcolsep}{1pt}
\centering
    \scalebox{.75}{
    \begin{tabular}{@{}l l c c c | cccccccccccc | c@{}}
	\toprule
	& Method & \llap{w/ }3D & PnP & Backbone & aero & bike & boat & bottle & bus & car & chair & table & mbike & sofa & train & tv & mean \\
	\midrule
	
	\parbox[t]{2mm}{\multirow{6}{*}{\rotatebox[origin=c]{90}{$\bm{\MedErr}\downarrow$}}} & Grabner et al.~\cite{Grabner20183DPE} & & \checkmark & ResNet-50 & 10.9 & {\bf 12.2} & 23.4 & 9.3 & 3.4 & 5.2 & 15.9 & 16.2 & 12.2 &  11.6 & 6.3 & 11.2 & 11.5 \\
	& StarMap~\cite{zhou2018starmap} & & \checkmark$^\dagger$ & ResNet-18 * & 10.1 & 14.5 & 30.3 & 9.1 & 3.1 & 6.5 & 11.0 & 23.7 & 14.1 & 11.1 & 7.4 & 13.0 & 12.8 \\
	& 3DPoseLite~\cite{Dani_2021_WACV} & \checkmark & & ResNet-18 &  -- & -- & -- & -- & -- & -- & -- & -- & -- & -- & -- & -- & 13.4 \\
	& PoseFromShape~\cite{Xiao2019PoseFS} & \checkmark & & ResNet-18 & 11.1 & 14.4 & 22.3 & 7.8 & 3.2 & 5.1 & 12.4 & 13.8 & 11.8 & {\bf 8.9} & {\bf 5.4} & {\bf 8.8} & 10.4 \\
	& PoseFromShape~\cite{Xiao2019PoseFS} & \checkmark & & ResNet-50 & 10.9 & 14.5 & 21.5 & 7.5 & 3.3 & 5.0 & 11.2 & 14.2 & 11.6 & 9.2 & 5.5 & 9.0 & 10.3 \\
    & PoseContrast (ours) & & & ResNet-50 & {\bf 10.0} & 13.6 & {\bf 18.3} & {\bf 7.2} & {\bf 2.8} & {\bf 4.6} & {\bf 9.8} & {\bf 9.2} & {\bf 11.5} & 11.0 & 5.6 & 11.6 & {\bf 9.6} \\
	\midrule
	
	\parbox[t]{2mm}{\multirow{6}{*}{\rotatebox[origin=c]{90}{$\bm{\AccThirty}\uparrow$}}} & Grabner et al.~\cite{Grabner20183DPE} & & \checkmark & ResNet-50 & 0.80 & 0.82 & 0.57 & 0.90 & {\bf 0.97} & 0.94 & 0.72 & 0.67 & 0.90 & 0.80 & {\bf 0.82} & {\bf 0.85} & 0.81 \\
	& StarMap~\cite{zhou2018starmap} & & \checkmark$^\dagger$ & ResNet-18 * & 0.82 & {\bf 0.86} & 0.50 & 0.92 & {\bf 0.97} & 0.92 & 0.79 & 0.62 & 0.88 & {\bf 0.92} & 0.77 & 0.83 & 0.82 \\
	& 3DPoseLite~\cite{Dani_2021_WACV} & \checkmark & & ResNet-18 & 0.80 & 0.82 & 0.58 & 0.93 & 0.96 & 0.92 & 0.77 & 0.57 & 0.88 & 0.82 & 0.80 & 0.79 & 0.80 \\
	& PoseFromShape~\cite{Xiao2019PoseFS} & \checkmark & & ResNet-18 & 0.83 & {\bf 0.86} & 0.60 & {\bf 0.95} & 0.96 & 0.91 & 0.79 & 0.67 & 0.85 & 0.85 & {\bf 0.82} & 0.82 & 0.83 \\
	& PoseFromShape~\cite{Xiao2019PoseFS} & \checkmark & & ResNet-50 & 0.83 & {\bf 0.86} & 0.61 & {\bf 0.95} & 0.96 & 0.92 & 0.80 & 0.67 & 0.84 & 0.82 & {\bf 0.82} & 0.83 & 0.83 \\
    & PoseContrast (ours) & & & ResNet-50 & {\bf 0.85} & 0.84 & {\bf 0.64} & 0.94 & {\bf 0.97} & {\bf 0.95} & {\bf 0.86} & {\bf 0.71} & {\bf 0.91} & 0.90 & {\bf 0.82} & {\bf 0.85} & {\bf 0.85} \\
	\bottomrule
	\end{tabular}
    }
    \vspace*{0.5mm}
    \caption{{\bf 3D Pose Estimation of Class-Agnostic Methods on Pascal3D+~\cite{xiang2014pascal3d} (all classes seen).}
    \yang{All methods are evaluated with ground-truth bounding boxes.}
    \yang{*\renaud{The authors} observe similar or worse performance with ResNet-50 \cite{zhou2018starmap}.} $^\dagger$StarMap actually obtains the rotation by solving for a similarity transformation between the image frame and world frame, weighting keypoint distances by the heatmap value.
    }
  \label{tab:Pascal3D}
  \vspace{2mm}
    \scalebox{.8}{
    \begin{tabular}{l l c c |cccccc|ccc|c c}
	\toprule
	& & & & 46 & 61 & 130 & 166 & 297 & 394 & 739 & 1092 & 2894 &  & 5818 \\
	& Method & \llap{w/ }3D & 2D Bbox & tool & misc & b-case & w-drobe & desk & bed & table & sofa & chair & mean & global \\
	\midrule
	\multirow{5}{*}{\rotatebox[origin=c]{90}{$\bm{\AccThirty}\uparrow$}} & 3DPoseLite~\cite{Dani_2021_WACV} & \checkmark & GT & \textbf{0.09} & 0.10 & 0.62 & 0.57 & 0.66 & 0.58 & 0.40 & 0.94 & 0.50 &  0.50 & 0.58 \\
	& PoseFromShape~\cite{Xiao2019PoseFS} * & \checkmark & GT & 0.07 & {\bf 0.28} & 0.71 & 0.65 & 0.71 & 0.54 & 0.53 & 0.94 & 0.79 & 0.58  & 0.75  \\
    & PoseContrast (ours) & & GT & \textbf{0.09} & 0.18 & {\bf 0.81} & {\bf 0.68} & {\bf 0.78} & {\bf 0.68} & {\bf 0.54} & {\bf 0.97} & {\bf 0.86} & {\bf 0.62} & {\bf 0.80} \raisebox{-2mm}{} \\
	\cline{2-15}
    & PoseFromShape~\cite{Xiao2019PoseFS}\raisebox{4mm}{} & \checkmark & pred & 0.07 & \textbf{0.23} & 0.68 & 0.55 & 0.71 & 0.51 & \textbf{0.53} & 0.93 & 0.77 & 0.55 & 0.73 \\
    & PoseContrast (ours) & & pred & \textbf{0.09} & 0.16 & \textbf{0.72} & \textbf{0.58} & \textbf{0.77} & \textbf{0.65} & \textbf{0.53} & {\bf 0.97} & \textbf{0.85} & \textbf{0.59} & \textbf{0.79} \\
	\bottomrule
	\end{tabular}
    }
    \vspace*{1.5mm}
    \caption{{\bf Cross-dataset 3D Pose Estimation of Class-Agnostic Methods on Pix3D~\cite{pix3d}.}
    The methods are trained on the Pascal3D+ \cite{xiang2014pascal3d}  train set and tested on Pix3D, where 6 classes are unseen (novel) and 3 classes are already seen (table, sofa, chair). As the classes are heavily unbalanced, we also report the global average (instance-wise rather than class-wise). We consider two kinds of input: ground-truth (GT) 2D object bounding box, and predicted (pred) bounding box by a class-agnostic Mask R-CNN detector. *3DPoseLite~\cite{Dani_2021_WACV} reports much worse figures for PoseFromShape~\cite{Xiao2019PoseFS} than what we got here with our own runs, probably due to a wrong experimental setting. 
    }
  \label{tab:Pix3D}
\vspace{-3mm}
\end{table*}

\paragraph{Datasets.\!\!\!}

We experiment with 3 commonly-used datasets for object pose estimation. Pascal3D+~\cite{xiang2014pascal3d} contains the 12 rigid classes of PASCAL VOC 2012~\cite{Everingham2009VOC}, with approximate poses. ObjectNet3D has slightly more accurate poses for 100 object classes. Pix3D~\cite{pix3d} features 9 classes, two of them (`tool' and `misc') not appearing in Pascal3D+ nor ObjectNet3D, with even more accurate poses.

\paragraph{Evaluation Metrics.}

Unless otherwise stated, ground-truth object bounding boxes are used in all experiments. We compute the most common metrics \cite{ViewpointsKeypoints2015,Su2015RenderFC}: $\AccThirty$ is the percentage of estimations with rotation error less than 30 degrees; $\MedErr$ is the median angular error in degrees. 



\subsection{Main Results}
\label{sec:expResults}

\paragraph{Upper Bound: Performance on Seen Classes.}

To check our performance before considering unseen classes, we first evaluate on seen classes. 
We follow the common protocol \cite{Grabner20183DPE,zhou2018starmap} to train our model on the train split of Pascal3D+ \cite{xiang2014pascal3d} and test it on the val split. 
Both train and val splits share the same 12 object classes.
In Table~\ref{tab:Pascal3D}, we compare with state-of-the-art class-agnostic object pose estimation methods\renaud{, using ground-truth bounding boxes. As we leverage on contrast-based features, we use the available MOCOv2 pre-trained ResNet-50 model~\cite{chen2020mocov2}. But no MOCO pre-trained ResNet-18 is available for comparison (see also Table \ref{tab:ablateFeat}).}

For most categories, our class-agnostic approach consistently outperforms other class-agnostic methods \cite{Grabner20183DPE,zhou2018starmap}, including those that leverage a 3D shape as additional input \cite{Xiao2019PoseFS,Dani_2021_WACV}. In particular, our direct pose estimation method achieves a clear improvement for the class `chair', which features a higher variety and geometric complexity than other classes. It suggests that keypoint-based methods as \cite{Grabner20183DPE,zhou2018starmap} may fail to capture detailed shape information for accurate 2D-3D correspondence prediction, while model-based methods as \cite{Xiao2019PoseFS,Dani_2021_WACV} do not construct powerful-enough embeddings despite their access to an actual 3D shape.

Overall, we achieve the best average performance in both metrics. In fact, we even outperform class-specific methods \cite{MahendranCVPRW2017,ViewpointsKeypoints2015,Mousavian20163DBB,Su2015RenderFC,deepdirectstat2018,Grabner20183DPE,MahendranBMVC2018} except one \cite{Liao2019SphericalRL}, that reaches {\MedErr} 9.2{\degree} and {\AccThirty} 88\%, while we get 9.6{\degree} and 85\%.

\begin{figure*}[!t]
    \centering

    \includegraphics[width=0.95\textwidth]{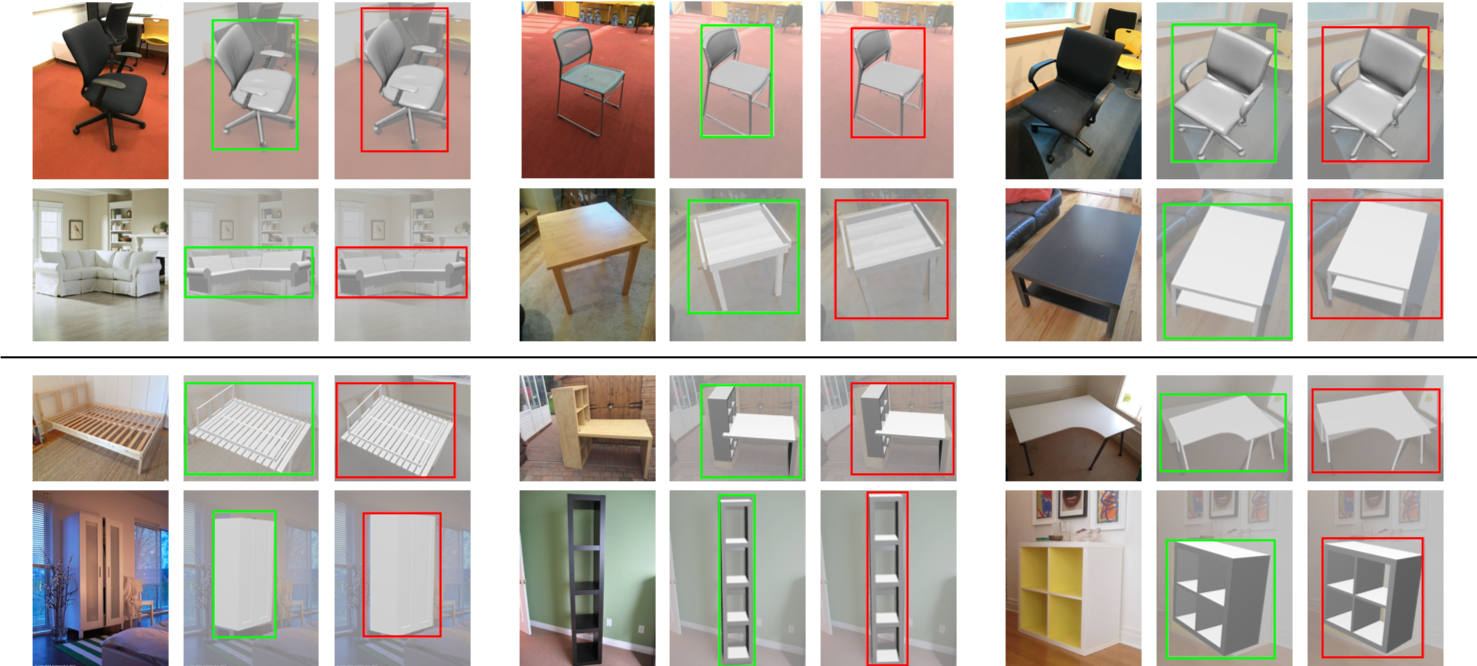}
    \caption{{\bf Qualitative Results on Pix3D.} For each sample, we first plot the original image, then we visualize the pose prediction obtained from the \textcolor{Green}{ground-truth bounding box} and the \textcolor{Red}{detected bounding box}, respectively. The top two rows show results for seen classes that intersect with training data in Pascal3D+ (`chair', `sofa', `table'), while the bottom two rows show results for novel classes.
    Note that the 3D CAD object models are only used here for pose visualization purpose; our approach does not rely on them for object pose prediction.
    }
    \label{fig:Pix3D}
    \vspace*{-3mm}
\end{figure*}

\paragraph{Stressing Class Agnosticism: Cross-Dataset Evaluation.}

To show our generalization ability, 
we  follow the recent protocol proposed in~\cite{Dani_2021_WACV} and conduct a cross-dataset object pose estimation. We train on the 12 classes of Pascal3D+ (that has approximate pose annotations) and test on the 9 classes of Pix3D (with accurate poses), where only 3 classes coincide with Pascal3D+. Hence, 6 classes are totally unseen while 3 are already seen.
Besides, methods that report cross-dataset results on Pix3D usually assume that ground-truth bounding boxes and 3D object models are given for testing~\cite{Xiao2019PoseFS,Dani_2021_WACV}. We compare here in that same setting (see below for using detected objects).
Results are in Table~\ref{tab:Pix3D}.
 
For the three seen classes (`table', `sofa', `chair'), our method outperforms all compared methods. It is consistent with results on Pascal3D+ (Table~\ref{tab:Pascal3D}), including for the difficult class `chair'. 
More interestingly, we achieve a significantly better performance for certain unseen classes, even though there is no prior knowledge of the testing objects for our network. As expected, it applies to unseen classes that share a similar shape and canonical frame as seen classes, e.g., `desk' and `table'. 
By sharing weights across different classes during training, our class-agnostic pose estimation network learns a direct mapping from image embeddings to 3D poses and can apply to unseen objects when they have a similar shape as the training objects. But when the target objects possess a geometry widely differing from the training ones, such as `tool' and `misc', our purely image-based method usually fails; PoseFromShape does a bit better because it leverages a shape model, but accuracy remains poor. 
Some failure cases of our method can be seen in Figure~\ref{fig:Pix3D_err}. 


\paragraph{Few-Shot Regime on ObjectNet3D.}

\begin{table}[!t]
\addtolength{\tabcolsep}{2pt}
    \centering
    \scalebox{.8}{
    \begin{tabular}{l c c c c}
    \toprule
    Method & Setting & w/ 3D & $\AccThirty \uparrow$ & $\MedErr \downarrow$ \\
    \midrule
    
    StarMap~\cite{zhou2018starmap}  & no-shot & & 0.44 & 55.8 \\
    PoseContrast (ours)  & no-shot & & \textbf{0.55} & \textbf{42.6} \\
    \hdashline
    \textcolor{mygray}{PoseFromShape}~\cite{Xiao2019PoseFS}  & \textcolor{mygray}{no-shot} & \textcolor{mygray}{\checkmark} &
    \textcolor{mygray}{0.62} & \textcolor{mygray}{42.0} \\
    \midrule
    MetaView~\cite{Tseng2019FewShotVE}  & 10-shot & & 0.48 & 43.4 \\
    PoseContrast (ours)  & 10-shot & & \textbf{0.60} & \textbf{38.7} \\
    \hdashline

    \textcolor{mygray}{FSDetView}~\cite{Xiao2020FewShotOD}  & \textcolor{mygray}{10-shot} & \textcolor{mygray}{\checkmark} & \textcolor{mygray}{0.63} & \textcolor{mygray}{32.1} \\
    \bottomrule
    \end{tabular}
    }
    \vspace{0.5mm}
    \caption{{\bf Few-Shot Object Pose Estimation on ObjectNet3D~\cite{xiang2016objectnet3d}\rlap.} 
    We report results on the 20 novel classes of ObjectNet3D as defined in \cite{zhou2018starmap,Tseng2019FewShotVE}. We compare both in no-shot and 10-shot settings. 
    }
    \label{tab:ObjectNet3D}
    \vspace*{-2.5mm}
\end{table}


We first follow the no-shot setting proposed in \cite{zhou2018starmap}: we train on 80 seen classes and test on 20 unseen (novel) classes, cf.\ Table~\ref{tab:ObjectNet3D} (top).
Compared to PoseFromShape~\cite{Xiao2019PoseFS}, both our approach and StarMap~\cite{zhou2018starmap} do not rely on 3D object models at test time, but exploit geometric similarities shared between seen and unseen classes. However, while StarMap struggles to predict precise 3D object coordinates and depth values, our simpler network achieve a higher performance.

We then evaluate in the 10-shot setting as in \cite{Tseng2019FewShotVE,Xiao2020FewShotOD}: the networks are first trained on the 80 seen classes, and then fine-tuned with a few labeled images from the 20 novel classes. 
Results are shown in Table~\ref{tab:ObjectNet3D} (bottom).
Compared to MetaView~\cite{Tseng2019FewShotVE}, that relies on class-specific keypoint prediction, we again find that our approach can obtain a better performance by sharing weights across all object classes.

In both settings, the best performing methods additionally use 3D object models~\cite{Xiao2019PoseFS,Xiao2020FewShotOD}. Such a prior knowledge of the geometry makes sense, especially for unseen objects with shapes widely different shapes from training classes.
Yet, our method nonetheless achieves promising results on these unseen classes, even compared to using a 3D model.

\vspace{-2mm}
\paragraph{Class-Agnostic Object Detection and Pose Estimation.}

To evaluate the coupling of generic object detection \emph{and} generic pose estimation, we train a Mask R-CNN with backbone ResNet-50 on COCO in a class-agnostic way, merging all classes into a single one, then apply it directly on Pix3D without fine-tuning. 
All 9 Pix3D classes can thus be detected by our network, whether they are in COCO or not.
To compare with other methods, we adopt the \rmf{well-established} 
metric $\mathit{Acc}_{D_{0.5}}$ \cite{Grabner2019GP2CGP}, that computes the percentage of objects for which the Intersection-over-Union (IoU) between the ground-truth and the predicted boxes is larger than $50\%$ \rmf{(ignoring false positives), to focus on objects of interest}.


Compared to class-specific methods that predict object localization together with their class \cite{wang20183d,Grabner2019GP2CGP}, our class-agnostic detector localizes objects without classifying them into categories, relying less on semantic information for prediction.
As shown in Figure~\ref{fig:DetectPix3D}, it provides a better detection accuracy and, more importantly, it enables the efficient detection of objects that are not included in COCO classes.

\definecolor{Bar1}{rgb}{0.47, 0.59, 1.00}
\definecolor{Bar2}{rgb}{0.59, 0.89, 0.47}
\definecolor{Bar3}{rgb}{1.00, 0.55, 0.51}

\begin{figure}[t]
\vspace*{-1mm}
    \centering \footnotesize
    \colorbox{Bar1}{\phantom{i}\phantom{i}} \cite{wang20183d} \quad \colorbox{Bar2}{\phantom{i}\phantom{i}} \cite{Grabner2019GP2CGP} \quad \colorbox{Bar3}{\phantom{i}\phantom{i}} Ours \\
    
    \includegraphics[width=0.85\columnwidth]{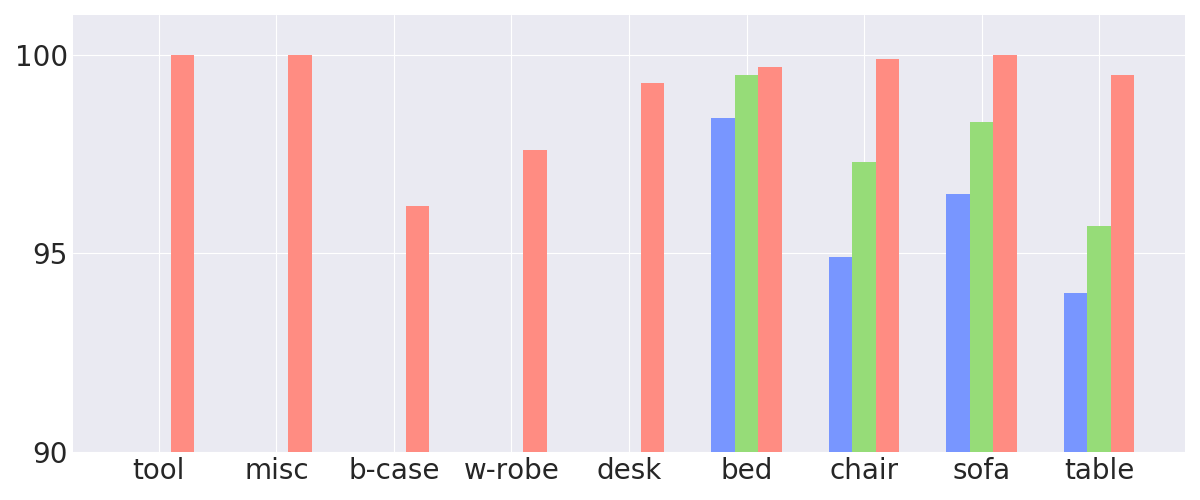}

    \caption{{\bf Object Detection on Pix3D.} Results are given in $\mathit{Acc}_{D_{0.5}}$ as defined in \cite{Grabner2019GP2CGP}. 
    We compare with two methods  \cite{wang20183d,Grabner2019GP2CGP} that train a class-specific Mask R-CNN on COCO, then fine-tune on a subset of Pix3D containing the same classes as COCO. In contrast, our agnostic Mask R-CNN is only trained on COCO and can generalize to classes not included in COCO.}
    \label{fig:DetectPix3D}
    \vspace*{-1mm}
\end{figure}


Qualitative results are shown in Figure~\ref{fig:Pix3D}. 
We find that both our object detector and our pose estimator can generalize to unseen objects (two bottom rows).
Quantitative results are given in Table~\ref{tab:Pix3D}. We observe that our object pose estimation, evaluated using predicted boxes, can outperform existing methods evaluated using ground-truth boxes only.
This promising results suggests it is possible to develop autonomous systems that perform class-agnostic object detection and pose estimation on unknown objects in the wild.

\subsection{Ablation Study}
\label{sec:expAblation}

\paragraph{Pre-trained Features.}
We initialize our image encoder network with MOCOv2~\cite{chen2020mocov2} to transfer rich features to the down-stream task of object pose estimation. Yet, other pre-trained features could be used \cite{He2020moco,chen2020simCLR,caron2020swav}, or learning from scratch. 
Table \ref{tab:ablateFeat} reports results with various initializations. 

Learning from scratch is suboptimal, probably due to the small dataset size, hence the relevance of using a pre-trained network. Convergence is also 5 times faster. Also, contrast-based pre-trained networks tend to perform best. \renaud{In comparison, \cite{Grabner20183DPE} also pre-trains on ImageNet while \cite{Xiao2019PoseFS} has similar results with or without ImageNet pre-training. \cite{zhou2018starmap} uses a ResNet-18 trained from scratch for its keypoint-based 2-stack hourglass network. Pre-training is not known for \cite{Dani_2021_WACV}.}



\begin{table}[t]
    \centering \addtolength{\tabcolsep}{-2.3pt}
    \scalebox{0.8}{
    \begin{tabular}{l l c c c c}
    \toprule
    Initialization & Method & \hspace{-6mm}Pre-train data\hspace{-1mm}  &  Epochs & $\AccThirty\uparrow$ & $\MedErr\downarrow$\hspace{-1mm} \\
    \midrule
    from scratch  & random & ---     & 15\rlap* & 0.76 & 12.8  \\
    from scratch  & random & ---     & 75 & 0.81 & 11.9  \\
    supervised      & classification & ImageNet & 15 & 0.83 & 10.7  \\
    unsupervised & SimCLR~\cite{chen2020simCLR} &  ImageNet & 15 & 0.83 & 11.0 \\
    unsupervised & SWAV~\cite{caron2020swav}   &  ImageNet & 15 & 0.84 & 10.2 \\
    unsupervised & MOCOv1~\cite{He2020moco} &  ImageNet & 15 & 0.84 & 10.3 \\
    unsupervised & MOCOv2~\cite{chen2020mocov2} &  ImageNet & 15 & \textbf{0.85} & \textbf{9.6}  \\
    \bottomrule
    \end{tabular}}
    \vspace{0mm}
    \caption{{\bf Network Initializations Evaluated on Pascal3D+.} 
    We compare different initializations, training until convergence (*except for the first line), showing the number of epochs required.
    }
    \vspace{-2mm}
    \label{tab:ablateFeat}
\end{table}


\begin{table}[t]
    \centering \setlength{\tabcolsep}{2pt}
    \scalebox{0.85}{
    \begin{tabular}{l @{}c@{\hspace*{0mm}} | c c | c c}
    \toprule
    & & \multicolumn{2}{c|}{Pascal3D+} & \multicolumn{2}{c}{Pix3D} \\
    Loss & \hspace*{-8mm}$\dist(\pose_i, \pose_{k^-}\!)$ & $\AccThirty$ & $\MedErr$ & $\AccThirty$ & $\MedErr$ \\
    \midrule
    $\loss_\ang$                        & N/A \hspace*{8mm} & 0.83 & 10.2 & 0.56 & 36.1 \\
    $\loss_\ang {+} \loss_\infoNCE$ & $1$ \hspace*{8mm} & 0.83 & 10.0 & 0.57 & 35.2 \\
    $\loss_\ang {+} \loss_\poseNCE$ & $(\Delta(\pose_i, \pose_{k^-}\!) / \pi)^{\frac{1}{2}}$ & 0.84 & 9.8 & 0.61 & 31.3 \\
    $\loss_\ang {+} \loss_\poseNCE$ & $(\Delta(\pose_i, \pose_{k^-}\!) / \pi)^{2}$ & \bf 0.85 & 10.0 & \bf 0.62 & 32.6 \\
    $\loss_\ang {+} \loss_\poseNCE$ & \hspace{2.8mm}$\hphantom(\Delta(\pose_i, \pose_{k^-}\!) / \pi \hphantom{^{1/2})}$ & \bf 0.85 & \bf 9.6 & \bf 0.62 & \bf 29.3 \\
    \bottomrule
    \end{tabular}}
    \vspace*{0mm}
    \caption{{\bf Adding a Contrastive Loss, Alternative Pose Distances.} 
    }
    \label{tab:ablateDist}
    \vspace*{-3.5mm}
\end{table}

\paragraph{Adding a Contrastive Loss.}

Table~\ref{tab:ablateDist} shows the relevance of adding a contrastive loss to the angle loss for pose estimation. However, adding the original InfoNCE loss only brings a minor improvement. A larger performance gap is obtained with our pose-aware contrastive loss of Eq.\,\eqref{eq:poseNCE}.



\paragraph{Alternative Pose Distances.}

Our contrastive loss relies on a distance between two poses $\dist(\pose_q, \pose_k)$, defined as the normalized rotation difference $\Delta(\pose_q, \pose_k) / \pi$. Table~\ref{tab:ablateDist} compares this definition to two variants: square and square root of this distance. All three perform better than the InfoNCE loss of Eq.\,\eqref{eq:infoNCE}, but the linear distance performs best.




\subsection{Discussion}

\begin{figure}
    \centering
    \includegraphics[width=\columnwidth]{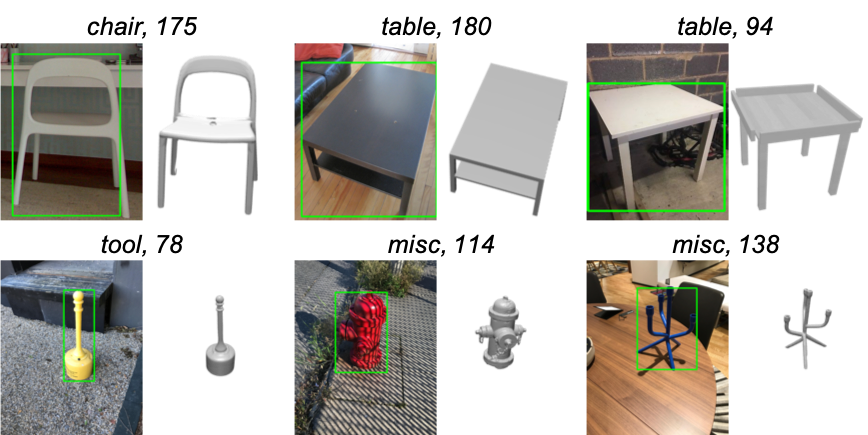}
    \caption{{\bf Visualization of Failure Cases.} We show input image crops and predicted object poses, with class name and prediction error displayed at the top. Common failures come from ambiguous appearances of symmetrical objects, or shapes out of distribution.
    }
    \label{fig:Pix3D_err}
    \vspace*{-3mm}
\end{figure}

\rmf{To understand where prediction errors come from, we show some common failure cases in Fig.\,\ref{fig:Pix3D_err}.
Many mistakes originate from symmetric objects, as their symmetry is neither modeled explicitly nor taken into account for metric evaluation (e.g., defining or measuring a table orientation by 180{\degree}). See a more detailed study in the supp.\,mat.}

\renaud{In fact, a few other works specifically treat symmetries \cite{Hodan2016Evaluation, Drost2017Introducing, Balntas2017PoseGR, corona2018pose, pitteri2019object}. It is largely orthogonal to our proposal and left for future work. Note that it concerns only about 10-15\% of the classes (e.g., table, bottle in Pascal3D+) and has little impact here as annotations generally assume the orientation with the smallest angle(s) w.r.t.\ the viewpoint.}

Our approach also fails on unseen objects with shapes differing completely from training ones, e.g., `tool' and `misc' of Pix3D. But it actually is a problem to all the RGB-only class-agnostic methods~\cite{zhou2018starmap,Grabner20183DPE}, not specifically to ours, as generalizing towards unseen objects mainly relies on similarities. 
Even shape-based methods~\cite{Xiao2019PoseFS,Dani_2021_WACV}, that exploit extra shape knowledge, nevertheless also struggle to get a good performance on these two classes. 

\section{Conclusion}

We presented a new class-agnostic object pose estimation approach based on a pose-aware contrastive learning. Our network is trained end-to-end, leveraging on existing unsupervised contrastive features. We empirically show on various benchmarks that our method constitutes a strong baseline for class-agnostic object pose estimation. We also pave the way to more practical applications by successfully combining it with a class-agnostic object detector. 


{\small
\bibliographystyle{ieee_fullname}
\bibliography{egbib}
}


\clearpage

\appendix
\setcounter{figure}{8}
\setcounter{table}{6}

This supplementary material to the ``PoseContrast'' paper (3DV 2021) provides:
\begin{itemize}
    \item[\ref{sec:supp_implem}.] Implementation and training details,
    \item[\ref{sec:supp_datasets}.] Dataset information,
    \item[\ref{sec:supp_param}.] Details on hyper-parameters,
    \item[\ref{sec:supp_tsne}.] Visualizations of the latent space,
    \item[\ref{sec:supp_classwise}.] Class-wise results on ObjectNet3D,
    \item[\ref{sec:supp_visual}.] Additional visual results.
    \item[\ref{sec:hist_azi}.] Histograms of azimuth prediction errors.
\end{itemize}
\emph{Note: reference numbers for citations used here are not the same as references used in the main paper; they correspond to the bibliography section at the end of this supplementary material.}

\section{Implementation and Training Details}\label{sec:supp_implem}

Our experiments are coded using PyTorch. Code will be made available upon publication.

\subsection{Training for 3D Pose Estimation}

We train our networks end-to-end using Adam optimizer with a batch size of 32 and an initial learning rate of 1e-4, which we divide by 10 at 80\% of the training phase. Unless otherwise stated, we train for 15 epochs, which takes less than 2 hours on a single V100-16G GPU. 

\subsection{Network and Training for Object Detection}

We use a class-agnostic Mask R-CNN~\cite{He2017MaskR} with a ResNet-50-FPN backbone~\cite{lin2017feature} as our instance segmentation network. The Mask~R-CNN is trained on COCO dataset~\cite{Lin2014MicrosoftCC}, which contains 80 classes and 115k training images. We use the open source repo of Mask~R-CNN and follow the training setting of \cite{He2017MaskR}, except that we adopt the class-agnostic architecture, where all 80 classes are merged into a single ``object'' category.

Our backbone network is initialized with weights pre-trained on ImageNet~\cite{Deng2009ImageNetAL}. During training, the shorter edge of images are resized to 800 pixels. Each GPU has 4 images and each image has 512 sampled RoIs, with a ratio of 1:3 of positives to negatives. We train our Mask R-CNN for 90k iterations. The learning rate is set to 0.02 at the beginning and is decreased by 10 at the 60k-th and 80k-th iteration. We use a weight decay of 0.0001 and momentum of 0.9. The entire training is carried out on 4 Nvidia RTX 2080Ti GPUs. During the training, mixed-precision training is used to reduce memory consumption and accelerate training.

\section{Datasets}\label{sec:supp_datasets}

We experimented with three commonly-used datasets for benchmarking object pose estimation in the wild. Table~\ref{tab:datasets} lists their main characteristics.

\begin{table}[h]
    \addtolength{\tabcolsep}{-2pt}
    \centering
    \scalebox{0.9}{
    \begin{tabular}{l c c c c}
    \hline
    Dataset     & year & \# classes & \# img train / val* & quality\\
    \hline
    Pascal3D+~\cite{xiang2014pascal3d} & 2014 & \hphantom{0}12 & 28,648 / \hphantom{0}2,113 & + \\
    ObjectNet3D~\cite{xiang2016objectnet3d} & 2016 & 100 & 52,048 / 34,375 & ++ \\
    Pix3D~\cite{pix3d}       & 2018 & \hphantom{00}9  & \quad\quad 0 / \hphantom{0}5,818 & +++\\
    \hline
    \end{tabular}}
    \vspace*{0.5mm}
    \caption{{\bf Experimented Datasets:} images of objects in the wild, with different qualities of pose annotation due to aligned shapes. *Only non-occluded and non-truncated objects, as done usually.
    }
    \label{tab:datasets}
\end{table}

While all these datasets feature a variety of objects and environments, Pascal3D+~\cite{xiang2014pascal3d} contains only the 12 rigid classes of PASCAL VOC 2012~\cite{Everingham2009VOC}, with approximate poses due to coarsely aligned 3D models at annotation time. ObjectNet3D distinguishes 100 classes in a subset of ImageNet~\cite{Deng2009ImageNetAL}, with more accurate poses as more and finer shapes were used for annotation. Recently, Pix3D~\cite{pix3d} proposes a smaller but more accurate dataset with pixel-level 2D-3D alignment using exact shapes; although it only features  9 classes, two of them (`tool' and `misc') do not appear in Pascal3D+ nor ObjectNet3D. We test all methods only on non-occluded and non-truncated objects, as other publications do.

\section{Hyper-parameters}\label{sec:supp_param}

We use parameters $\lambda\,{=}\,1$, $\kappa\,{=}\,1$, $\tau\,{=}\,0.5$, and the transformation rotation $\phi$ varies in $[-15\degree,15\degree]$, i.e., $[-\frac{\pi}{24},\frac{\pi}{24}]$. 

\begin{figure}[t]
    \centering
    \includegraphics[width=1.0\columnwidth]{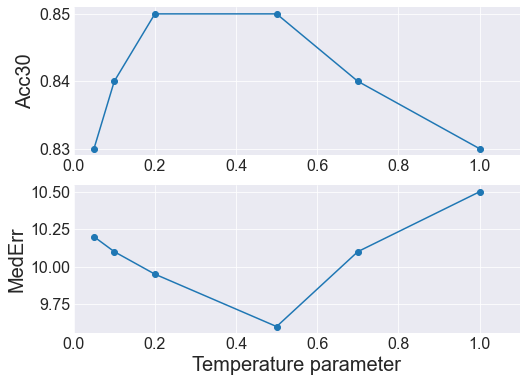}
    \caption{{\bf Parameter Study of Temperature $\tau$ in $\loss_\mathrm{poseNCE}$.} We report the performance on the dataset Pascal3D+~\cite{xiang2014pascal3d} with 30-degree accuracy ($\AccThirty\uparrow$) and median error ($\MedErr\downarrow$).}
    \label{fig:AblationTau}
    \vspace{-6mm}
\end{figure}

Figure~\ref{fig:AblationTau} shows the influence of temperature parameter $\tau$ in the proposed pose-aware contrastive loss $\loss_\mathrm{poseNCE}$. By varying this parameter from 0.05 to 1.0, we obtain the best performance on Pascal3D+ when $\tau=0.5$. While training without this pose-aware contrastive loss can still reach an overall accuracy at 0.83 and an overall median error at 10.2, we note that the performance can be improved using the propose loss $\loss_\mathrm{poseNCE}$ with a temperature parameter between 0.2 and 0.6, which is quite robust.

\section{Visualizations of the Latent Space}\label{sec:supp_tsne}

\begin{figure*}[!p]
    \centering
    \hspace{-6pt}\begin{tabular}{@{}cc@{}}
        Pascal3D~\cite{xiang2014pascal3d} & Pix3D~\cite{pix3d} \\[2mm]
        \includegraphics[width=0.49\linewidth,trim=105 52 112 85, clip]{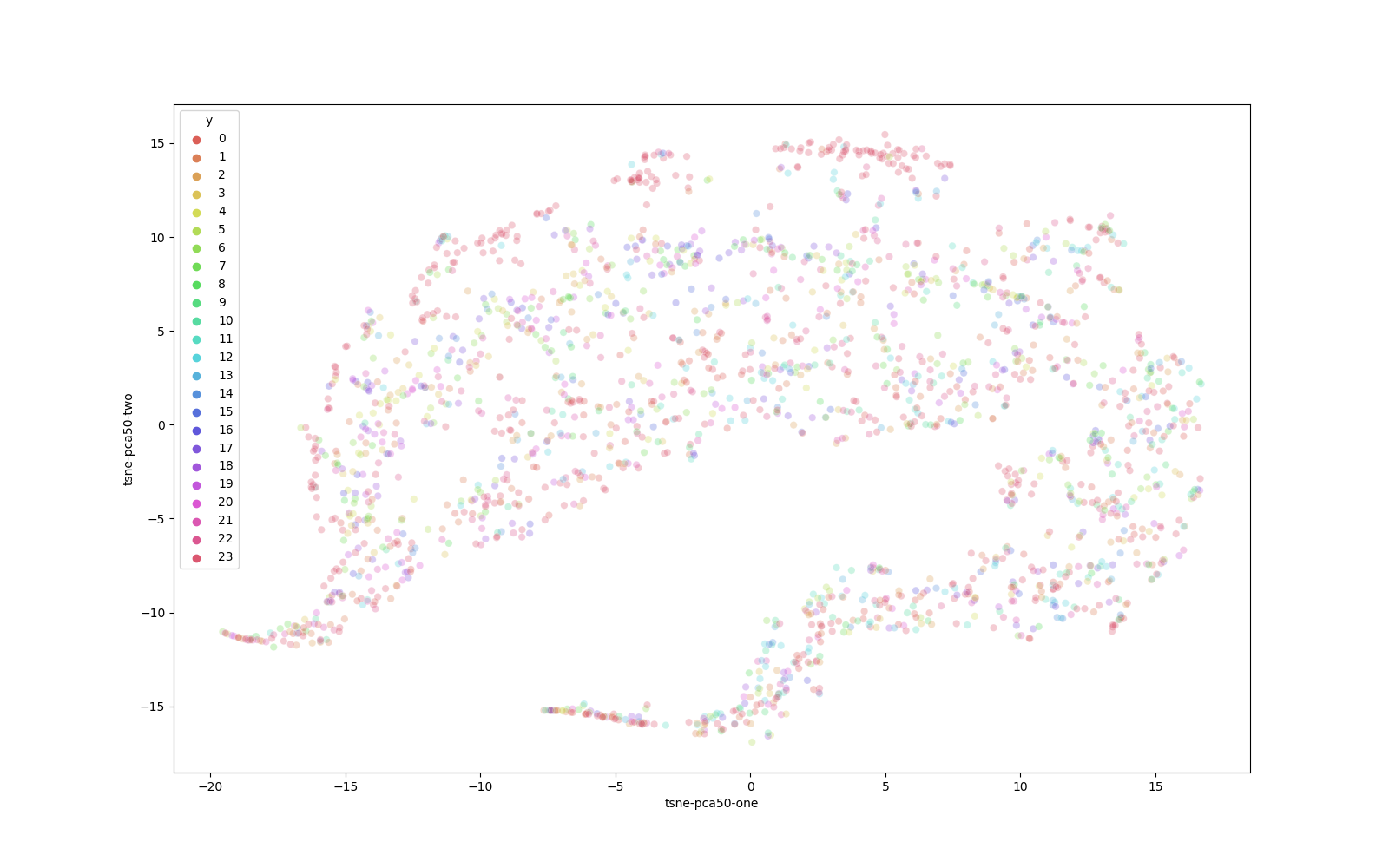} & 
        \includegraphics[width=0.49\linewidth,trim=105 52 112 85, clip]{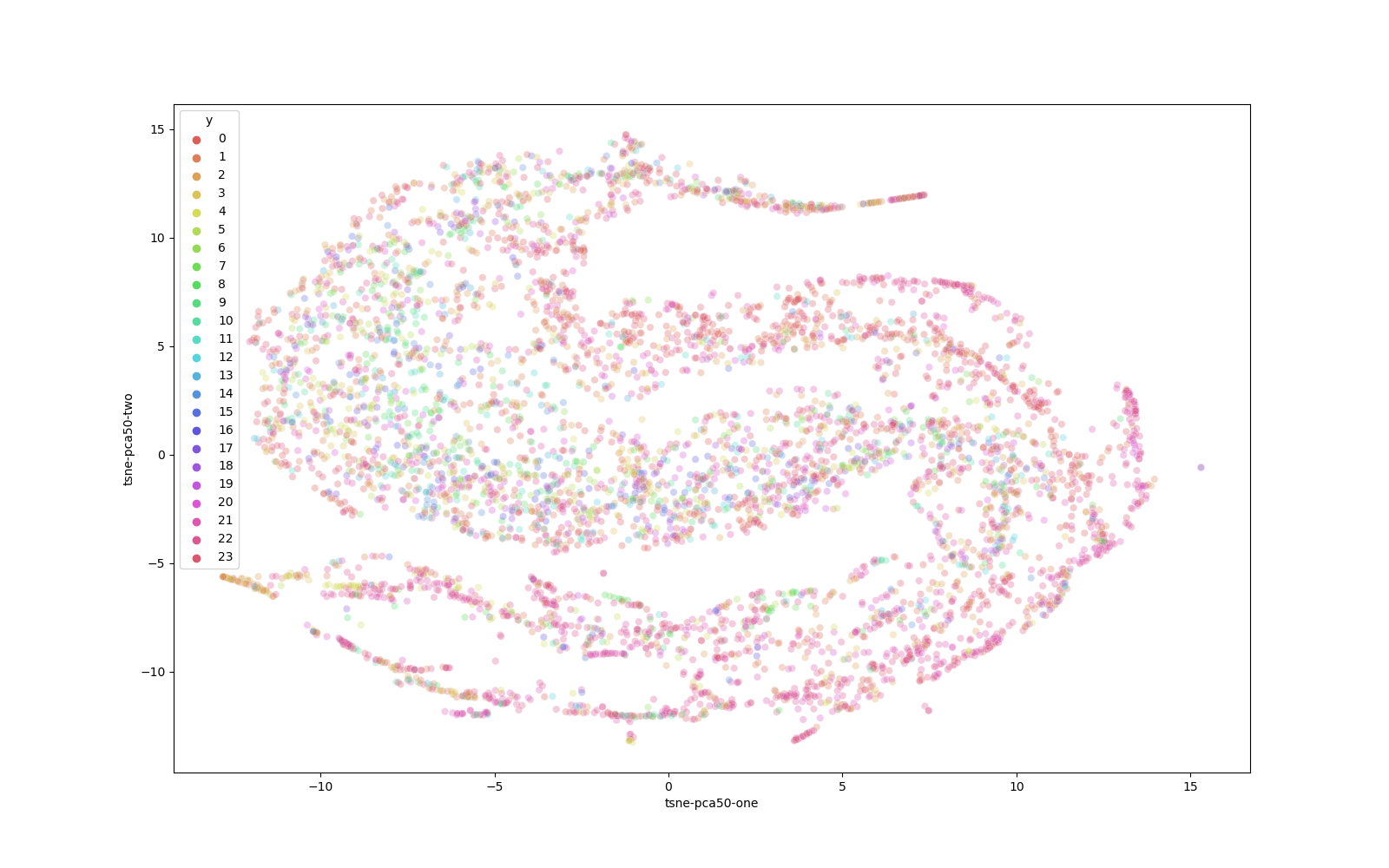}
        \\
        \small (a) Random & \small (b) Random
        \\[3mm]
        \includegraphics[width=0.49\linewidth,trim=105 52 112 85, clip]{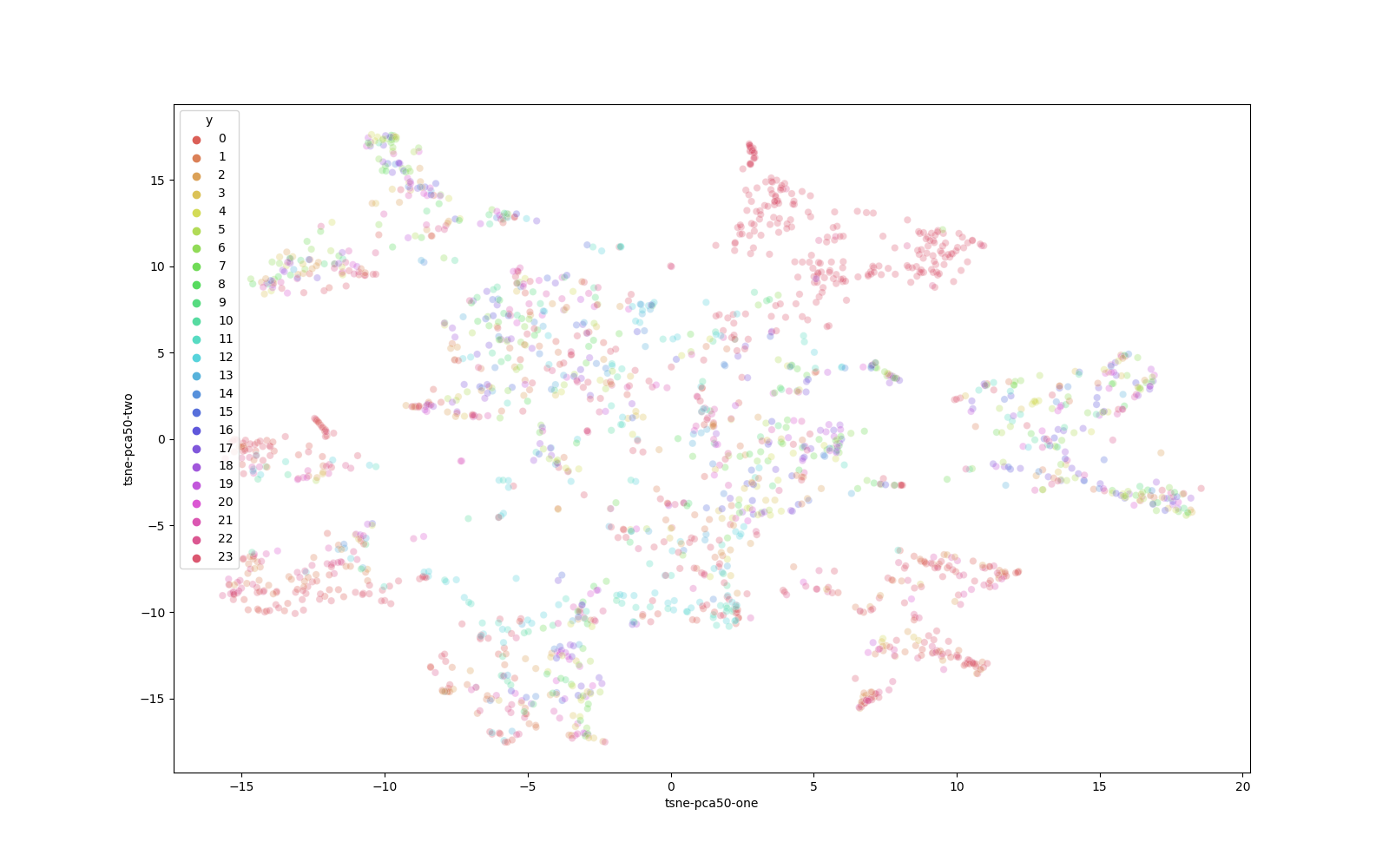} & \includegraphics[width=0.49\linewidth,trim=105 52 112 85, clip]{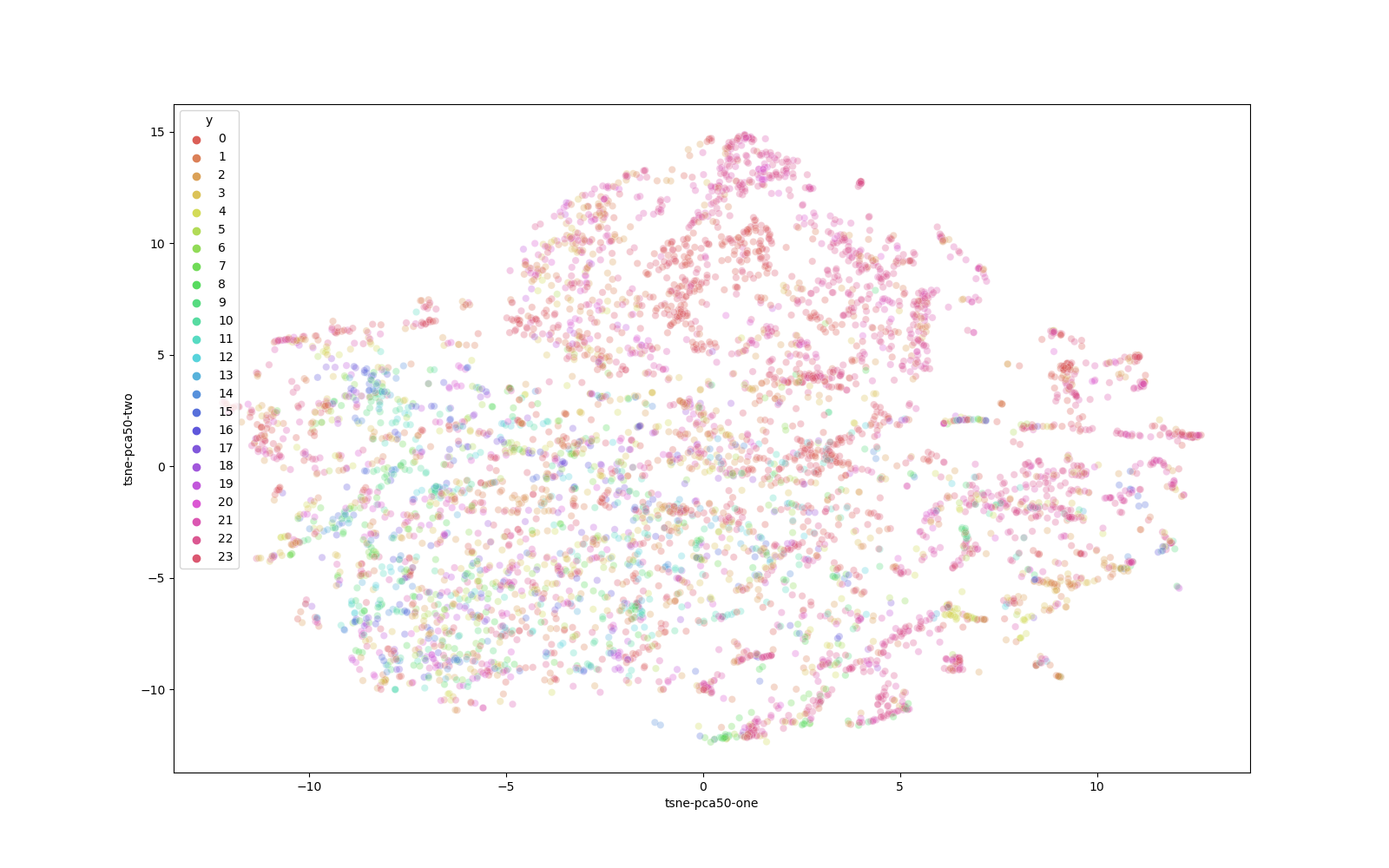}
        \\
        \small (c) MOCOv2 & \small (d) MOCOv2
        \\[3mm]
        \includegraphics[width=0.49\linewidth,trim=105 52 112 85, clip]{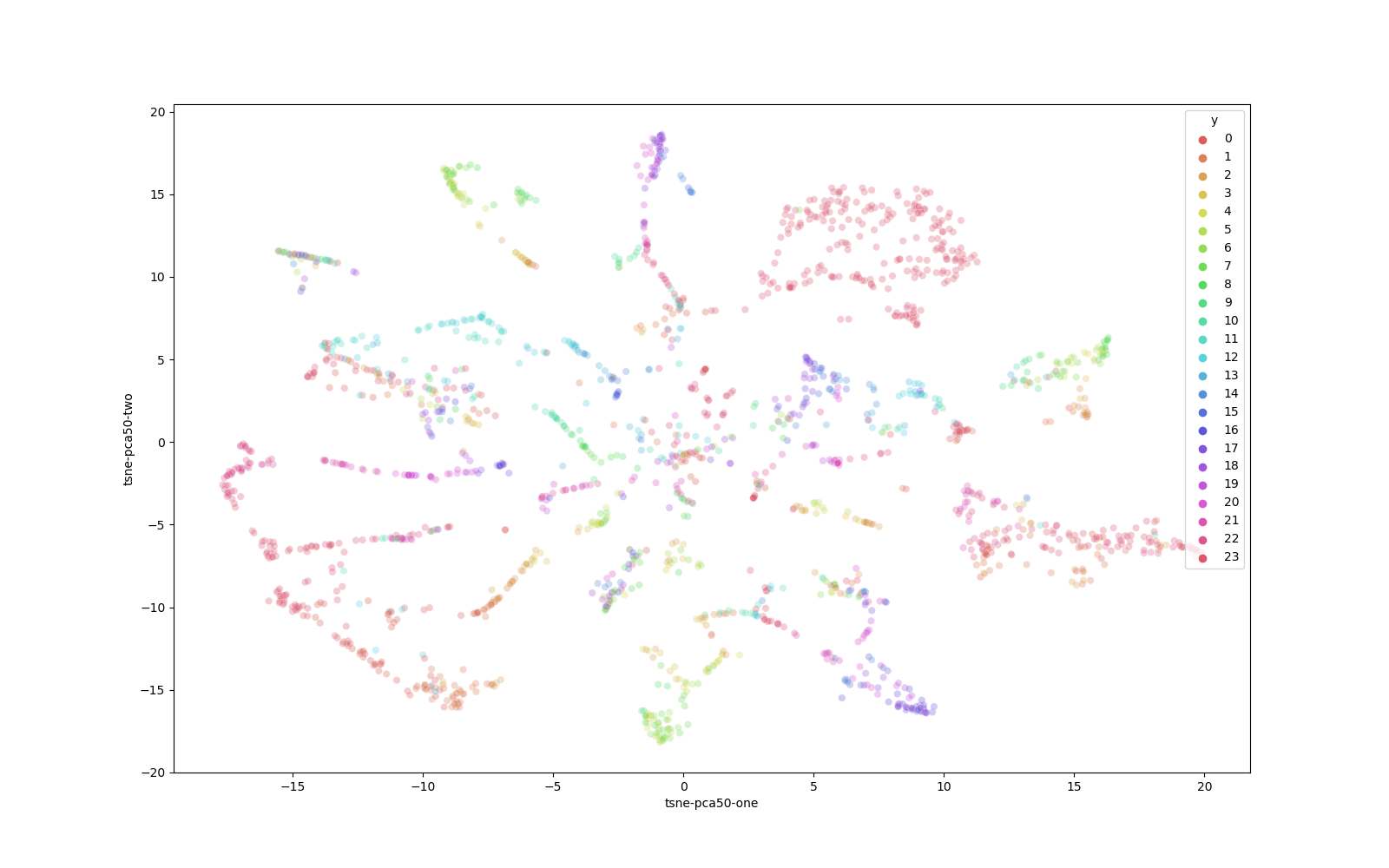} &
        \includegraphics[width=0.49\linewidth,trim=105 52 112 85, clip]{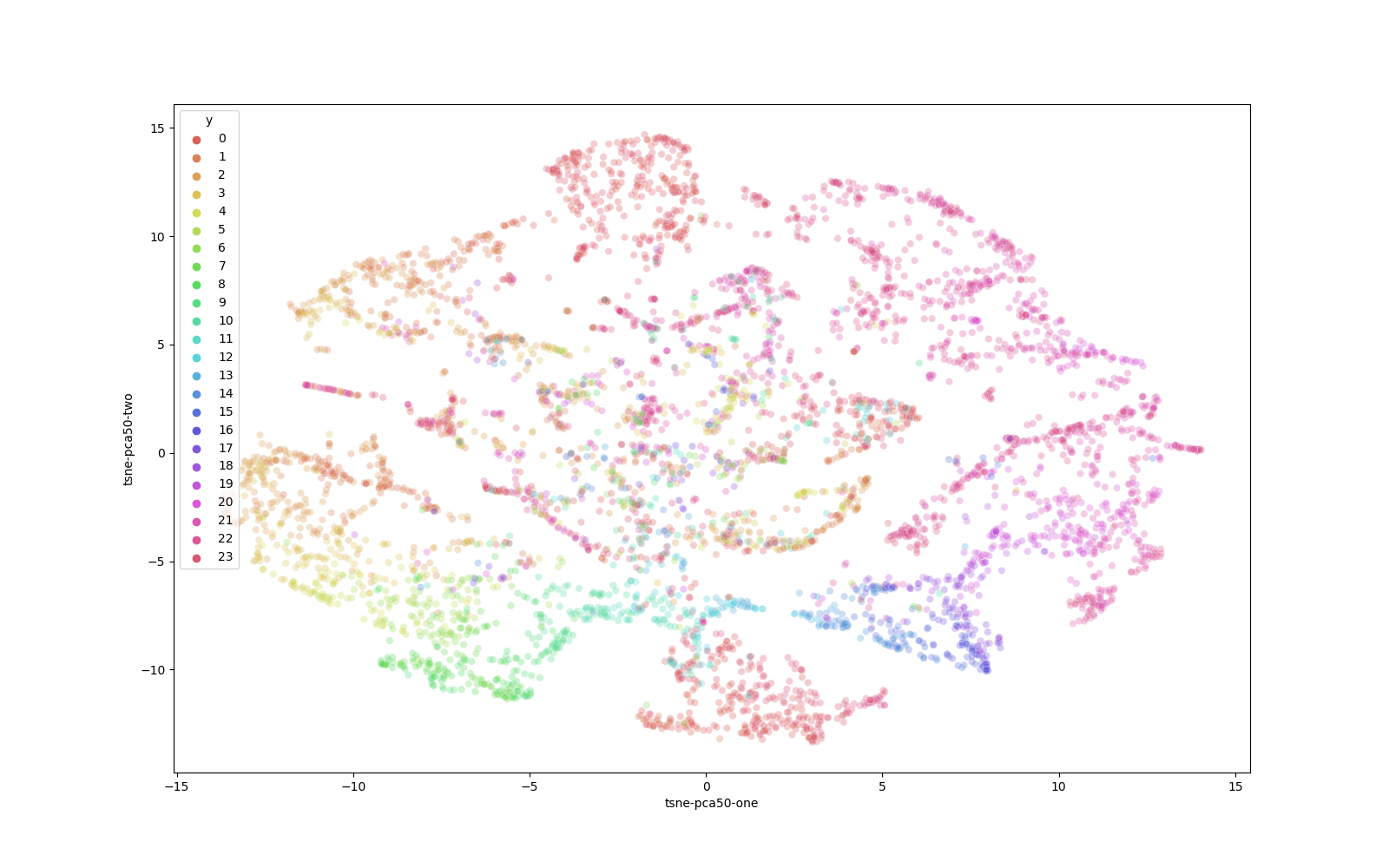}
        \\
        \small (e) PoseContrast & \small (f) PoseContrast
        \\[3mm]
    \end{tabular}
    \caption{{\bf Feature Visualization.} 
    We visualize image features from the val set of Pascal3D+~\cite{xiang2014pascal3d} (left) and Pix3D~\cite{pix3d} (right) by t-SNE (preceded by PCA) 
    for three different ResNet-50 backbones: (a,b)~randomly initialized network (top); (c,d)~network pre-trained on ImageNet by MOCOv2~\cite{chen2020mocov2} (middle); and (e,f)~network trained on Pascal3D+ with PoseContrast (bottom). We divide the 360 degrees of azimuth angle into 24 bins of 15{\degree} and use one color for each bin. The figure is better viewed in color with zoom-in.}
    \label{fig:tSNE}
\end{figure*}

To better understand the effect of contrastive learning, we use t-SNE to visualize the features obtained by different backbones. Results are presented in Figure~\ref{fig:tSNE}.

The features are extracted from val images of Pascal3D+. Considering the fact that the distributions of elevation and inplane-rotation are highly centered around a specific value compared to that of azimuth, we split the visualized features into different clusters, with each cluster corresponding to objects with similar azimuth angles. More specifically, we divide the 360 degrees of azimuth angle into 24 bins, and objects within the same azimuth bin are shown by the same color.

As seen in Figure~\ref{fig:tSNE} (left), the features extracted using a randomly-initialized network are more or less uniformly distributed across different locations in the latent space, and regardless of their 3D poses. On the contrary, the features extracted using networks trained with contrastive learning (MOCOv2 and PoseContrast) tend to form clusters, where each cluster groups objects with similar azimuth angles. Arguably, feature clusters are less spread with PoseContrast, compared to MOCOv2, and actual azimuths are more consistent within clusters.

When doing the same kind of visualization on Pix3D, as shown in Figure~\ref{fig:tSNE} (right), we observe more or less the same kind of distribution for the random initialization. However, MOCOv2 has a harder time clustering the features of Pix3D images, including regarding pose. Yet, PoseContrast manages to produce clusters, and with a better pose consistency.

\begin{table*}[t]
\addtolength{\tabcolsep}{-3.5pt}
\centering
    \scalebox{0.8}{
    \begin{tabular}{c c c c c c c c @{~}|| c c}
    \toprule
    \multicolumn{8}{c||}{Base classes} & \multicolumn{2}{c}{Novel classes} \\
    \midrule
    aeroplane & ashtray & backpack & basket & bench & bicycle & blackboard & boat & bed & bookshelf \\
    bottle & bucket & bus & cabinet & camera & can & cap & car & calculator & cellphone \\
    chair & clock & coffee\_maker & comb & cup & desk\_lamp & diningtable & dishwasher & computer & door \\
    eraser & eyeglass & fan & faucet & file\_extinguisher & fish\_tank & flashlight & fork & filling\_cabinet & guitar \\
    hair\_dryer & hammer & headphone & helmet & jar & kettle & key & keyboard & iron & knife \\
    laptop & lighter & mailbox & microphone & motorbike & mouse & paintbrush & pan & microwave & pen \\
    pencil & piano & pillow & plate & printer & racket & refrigerator & remote\_control & pot & rifle \\
    road\_pole & satellite\_dish & scissors & screwdriver & shovel & sign & skate & skateboard & shoe & slipper \\
    sofa & speaker & spoon & stapler & suitcase & teapot & telephone & toaster & stove & toilet \\
    toothbrush & train & train\_bin & trophy & tvmonitor & vending\_machine & washing\_machine & watch & tub & wheelchair \\
    \bottomrule
    \end{tabular}}
    \vspace*{2mm}
    \caption{{\bf Dataset split of ObjectNet3D~\cite{xiang2016objectnet3d}:} 80 base classes (left) and 20 novel classes (right). Some novel classes share similar geometries and canonical frames as base classes, e.g., `door'/`blackboard', `filling\_cabinet'/`cabinet', `wheelchair'/'chair'.}
    \label{tab:ClassSplit}
    \vspace{9mm}
	\scalebox{0.85}{
	\begin{tabular}{l l@{\hspace*{10mm}} c@{\hspace*{8mm}} c@{\hspace*{8mm}} c@{\hspace*{8mm}} c@{\hspace*{8mm}} c@{\hspace*{8mm}} c@{\hspace*{8mm}} c}
	\toprule
	&Method \ \ \emph{\small Acc30($\uparrow$) / MedErr($\downarrow$)}\hspace{-10mm}
	& bed & bookshelf & calculator & cellphone & computer & door & \hspace{-4mm}filling\_cabinet\hspace{-0mm} \\ 
	\midrule
	\parbox[t]{4mm}{\multirow{3}{*}{\rotatebox[origin=c]{90}{no-shot}}} &
	StarMap~\cite{zhou2018starmap} & 0.37 / 45.1 & 0.69 / 18.5 & 0.19 / 61.8 & 0.51 / 29.8 & 0.74 / 15.6 & -- / -- & 0.78 / 14.1 \\
	&PoseContrast (ours) & {0.62} / {17.4} & {0.89} / {~6.7} & {0.65} / {17.7} & {0.57} / {15.8} & {0.85} / {14.5} & {0.91} / {~2.7} & {0.88} / {10.4} \\
	\cdashline{2-9}
	&\textcolor{mygray}{PoseFromShape~\cite{Xiao2019PoseFS}} & \textcolor{mygray}{{0.65} / {15.7}} & \textcolor{mygray}{{0.90} / {~6.9}} & \textcolor{mygray}{{0.88} / {12.0}} & \textcolor{mygray}{{0.65} / {10.5}} & \textcolor{mygray}{{0.84} / {11.2}} & \textcolor{mygray}{{0.93} / {~2.3}} & \textcolor{mygray}{{0.84} / {12.7}} \\
	\midrule
	\parbox[t]{2mm}{\multirow{4}{*}{\rotatebox[origin=c]{90}{10-shot}}} &
	StarMap*~\cite{zhou2018starmap} & 0.32 / 42.2 & 0.76 / 15.7 & 0.58 / 26.8 & 0.59 / 22.2 & 0.69 / 19.2 & -- / -- & 0.76 / 15.5 \\
	& MetaView~\cite{Tseng2019FewShotVE} & 0.36 / 37.5 & 0.76 / 17.2 & 0.92 / 12.3 & 0.58 / 25.1 & 0.70 / 22.2 & -- / -- & 0.66 / 22.9 \\
	& PoseContrast (ours) & {0.67} / {13.9} & {0.90} / {~7.0} & {0.85} / {11.0} & {0.58} / {15.2} & {0.85} / {10.9} & {0.91} / {~2.5} & {0.89} / {~8.4}  \\
	\cdashline{2-9}
	& \textcolor{mygray}{FSDetView~\cite{Xiao2020FewShotOD}} & \textcolor{mygray}{{0.64} / {14.7}} & \textcolor{mygray}{{0.89} / {~8.3}} & \textcolor{mygray}{{0.90} / {~8.3}} & \textcolor{mygray}{{0.63} / {12.7}} & \textcolor{mygray}{{0.84} / {10.5}} & \textcolor{mygray}{{0.90} / {~0.9}} & \textcolor{mygray}{{0.84} / {10.5}} \\ \midrule \midrule
	
	& Method  \ \ \emph{\small Acc30($\uparrow$) / MedErr($\downarrow$)}\hspace{-10mm}
	& guitar & iron & knife & microwave & pen & pot & rifle \\
	\midrule
	\parbox[t]{2mm}{\multirow{3}{*}{\rotatebox[origin=c]{90}{no-shot}}} &
	StarMap~\cite{zhou2018starmap} & 0.64 / 20.4 & 0.02 / 142 & 0.08 / 136 & 0.89 / 12.2 & -- / -- & 0.50 / 30.0 & 0.00 / 104 \\
	&PoseContrast (ours) & {0.73} / {14.4} & {0.03} / {124} & {0.25} / {122} & {0.93} / {~7.5} & {0.45} / {39.8} & {0.76} / {~9.2} & {0.00} / {102} \\
    \cdashline{2-9}
	&\textcolor{mygray}{PoseFromShape~\cite{Xiao2019PoseFS}} & \textcolor{mygray}{{0.67} / {20.8}} & \textcolor{mygray}{{0.02} / {145}} & \textcolor{mygray}{{0.29} / {138}} & \textcolor{mygray}{{0.94} / {~7.7}} & \textcolor{mygray}{{0.46} / {37.3}} & \textcolor{mygray}{{0.79} / {13.2}} & \textcolor{mygray}{{0.15} / {110}} \\
	\midrule
	
	\parbox[t]{2mm}{\multirow{4}{*}{\rotatebox[origin=c]{90}{10-shot}}} &
	StarMap*~\cite{zhou2018starmap} & 0.59 / 21.5 & 0.00 / 136 & 0.08 / 117 & 0.82 / 17.3 &-- / --& 0.51 / 28.2 & 0.01 / 100 \\
	&MetaView~\cite{Tseng2019FewShotVE} & {0.63} / {24.0} & 0.20 / {76.9} & 0.05 / {97.9} & 0.77 / 17.9 &-- / --& 0.49 / 31.6 & ~0.21 / {80.9} \\
	&PoseContrast (ours) & {0.73} / {14.7} & {0.03} / {126} & {0.23} / {116} & {0.94} / {~6.9} & {0.45} / {41.3} & {0.78} / {10.6} & {0.04} / {90.4}  \\
    \cdashline{2-9}
	&\textcolor{mygray}{FSDetView~\cite{Xiao2020FewShotOD}} & \textcolor{mygray}{{0.72} / {17.1}} & \textcolor{mygray}{{0.37} / {57.7}} & \textcolor{mygray}{{0.26} / {139}} & \textcolor{mygray}{{0.94} / {~7.3}} & \textcolor{mygray}{{0.45} / {44.0}} & \textcolor{mygray}{{0.74} / {12.3}} & \textcolor{mygray}{{0.29} / {88.4}} \\ \midrule \midrule
	
	& Method  \ \ \emph{\small Acc30($\uparrow$) / MedErr($\downarrow$)}\hspace{-10mm}
	& shoe & slipper & stove & toilet & tub & wheelchair & \cellcolor{HL} TOTAL \\
	\midrule
	\parbox[t]{2mm}{\multirow{3}{*}{\rotatebox[origin=c]{90}{no-shot}}} &
	StarMap~\cite{zhou2018starmap} & -- / -- & 0.11 / 146 & 0.82 / 12.0 & 0.43 / 35.8 & 0.49 / 31.8 & 0.14 / 93.8 & \cellcolor{HL}{ 0.44 / 55.8 } \\
	&PoseContrast (ours) & {0.23} / {58.9} & {0.25} / {138} & {0.91} / {12.0} & {0.43} / {30.8} & {0.53} / {24.0} & {0.42} / {43.4} & \cellcolor{HL}{ {0.56} / {42.6} } \\
	\cdashline{2-9}
	&\textcolor{mygray}{PoseFromShape~\cite{Xiao2019PoseFS}} & \textcolor{mygray}{{0.54} / {28.2}} & \textcolor{mygray}{{0.32} / {158}} & \textcolor{mygray}{{0.89} / {10.1}} & \textcolor{mygray}{{0.61} / {21.8}} & \textcolor{mygray}{{0.68} / {17.8}} & \textcolor{mygray}{{0.39} / {57.4}} & \cellcolor{HL}{ \textcolor{mygray}{{0.62} / {42.0}} } \\
	\midrule
	
	\parbox[t]{2mm}{\multirow{4}{*}{\rotatebox[origin=c]{90}{10-shot}}} &
	StarMap*~\cite{zhou2018starmap} &-- / --& 0.15 / 128 & 0.83 / 15.6 & 0.39 / 35.5 & 0.41 / 38.5 & 0.24 / 71.5 & \cellcolor{HL}0.46 / 50.0 \\
	&MetaView~\cite{Tseng2019FewShotVE} &-- / --& 0.07 / 115 & 0.74 / 21.7 & 0.50 / 32.0 & 0.29 / 46.5 & 0.27 / 55.8 & \cellcolor{HL}0.48 / 43.4 \\
	&PoseContrast (ours) & {0.24} / {56.7} & {0.23} / {155} & {0.92} / {~8.1} & {0.64} / {22.2} & {0.55} / {18.6} & {0.45} / {36.7} & \cellcolor{HL}{ {0.60} / {38.7} } \\
	\cdashline{2-9}
	&\textcolor{mygray}{FSDetView~\cite{Xiao2020FewShotOD}} & \textcolor{mygray}{{0.51} / {29.4}} & \textcolor{mygray}{{0.25} / {96.4}} & \textcolor{mygray}{{0.92} / {~9.4}} & \textcolor{mygray}{{0.69} / {17.4}} & \textcolor{mygray}{{0.66} / {15.1}} & \textcolor{mygray}{{0.36} / {64.3}} & \cellcolor{HL}\textcolor{mygray}{{0.63} / {32.1}} \\ \bottomrule
	\end{tabular}}
	\vspace*{2mm}
	\caption{{\bf Few-shot viewpoint estimation on ObjectNet3D~\cite{xiang2016objectnet3d}.} 
	All models are trained and evaluated on ObjectNet3D. For each method, we report Acc30($\uparrow$) / MedErr($\downarrow$) on the same 20 novel classes of ObjectNet3D, while the remaining 80 classes are used as base classes. *StarMap network trained with MAML~\cite{Finn2017ModelAgnosticMF} for few-shot viewpoint estimation, with numbers reported in \cite{Tseng2019FewShotVE}. Methods additionally using 3D object models are shown in \textcolor{mygray}{gray}. }
	\vspace*{2mm}
	\label{tab:FewShotCls}
\end{table*}

\section{Class-Wise Results on ObjectNet3D}\label{sec:supp_classwise}

In this section, we present the class-wise quantitative results on the dataset ObjectNet3D~\cite{xiang2016objectnet3d}. As detailed in Table~\ref{tab:ClassSplit}, we follow previous work~\cite{zhou2018starmap,Tseng2019FewShotVE} and split the 100 object classes of ObjectNet3D into 80 base classes and 20 novel classes. We conduct both no-shot and 10-shot viewpoint estimation, as described in the paper, and report the performance on each novel classe in Table~\ref{tab:FewShotCls}.

We see that our method, that only relies on RGB images, significantly outperforms all other RGB-based methods~\cite{zhou2018starmap,Tseng2019FewShotVE} on both tasks and both metrics.

The best overall performance is however achieved by methods that additionally make use of 3D models. It was expected since these methods are designed to extract information regarding geometry and canonical frame from the aligned 3D object models. In fact, the performance of methods using CAD models can somehow be regarded as an upper bound with respect to RGB-based methods. Nevertheless, we outperform CAD-model-based methods on a few classes, e.g., `filling\_cabinet', `guitar' and  `wheelchair'. Besides, the relative gap between our method and these CAD-model-based methods is mostly due to a few classes, such as `rifle', `iron' or `shoe', for which base classes offer limited help in terms of geometrical cue or canonical frame.  In fact, if we put aside `iron' and `shoe', our method is on par with FSDetView~\cite{Xiao2020FewShotOD} on 10-shot viewpoint estimation, despite not using any extra shape information.


Moreover, our class-agnostic network directly estimates the viewpoint from image embeddings, without relying on any keypoint prediction. This direct estimation network thus can predict the viewpoint for all classes, while keypoint-based methods struggle to get a reasonable prediction for certain classes, e.g., `door', `pen', and `shoe'.

\section{Additional Visual Results}\label{sec:supp_visual}

\begin{figure*}
    \centering
    \includegraphics[width=0.95\textwidth]{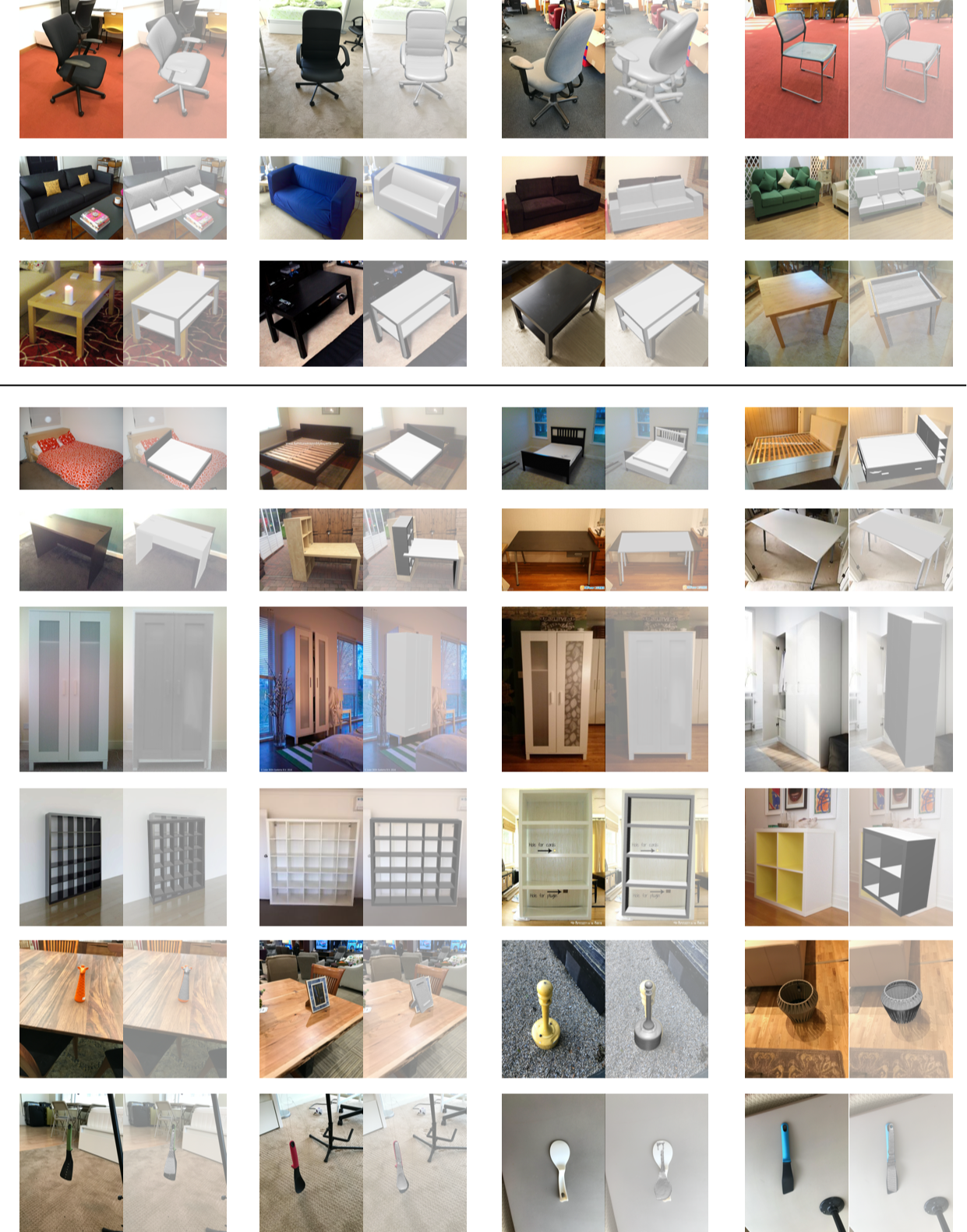}
    \vspace*{1mm}
    \caption{{\bf Additional qualitative results on the 9 object classes of Pix3D~\cite{pix3d}.} The network is trained on the 12 object classes of Pascal3D+ and directly tested on Pix3D. From top to bottom: `chair', `sofa', `table', `bed', `desk', `wardrobe', `bookcase', `misc', and `tool'. 3D object models here are only used to visualize the pose.}
    \label{fig:Pix3D_vis_cls}
\end{figure*}

We present in Figure~\ref{fig:Pix3D_vis_cls} some additional visual results of our class-agnostic method on the cross-dataset 3D pose estimation, training on Pascal3D+ and testing on Pix3D~\cite{pix3d}.

\section{Histograms of Azimuth Prediction Errors}\label{sec:hist_azi}

\begin{figure*}
    \centering \hspace{-6pt}
    \begin{tabular}{@{}ccc@{}}
    
    \small aeroplane & \small bicycle & \small boat \\
    \includegraphics[width=0.30\linewidth]{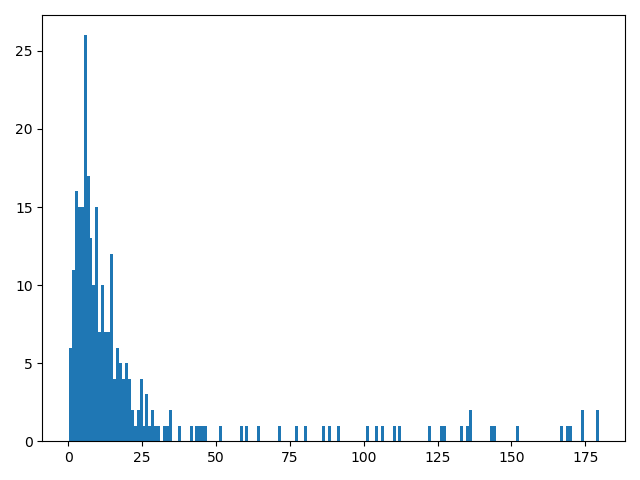} & 
    \includegraphics[width=0.30\linewidth]{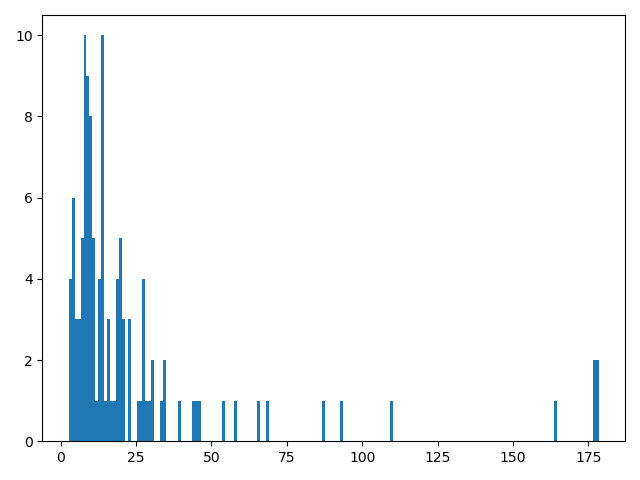} & 
    \includegraphics[width=0.30\linewidth]{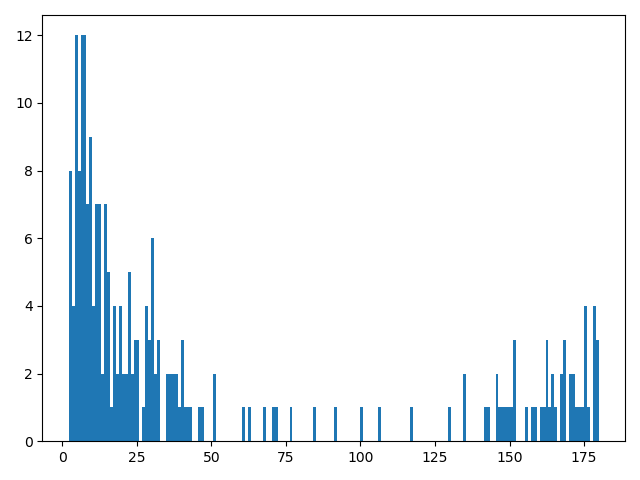} 
    \vspace{6mm} \\
    \small bottle & \small bus & \small car \\
    \includegraphics[width=0.30\linewidth]{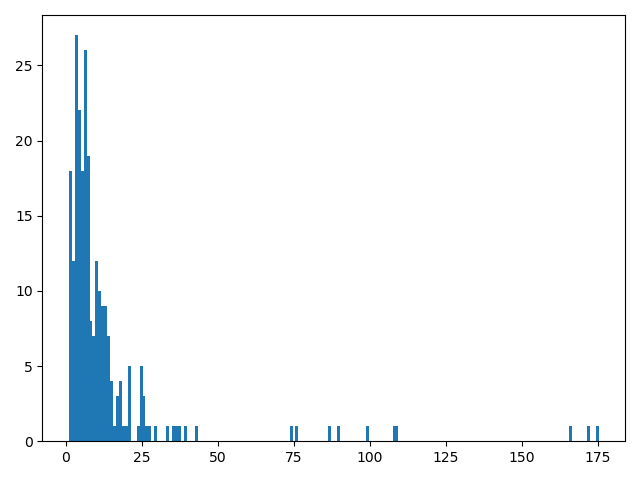} & 
    \includegraphics[width=0.30\linewidth]{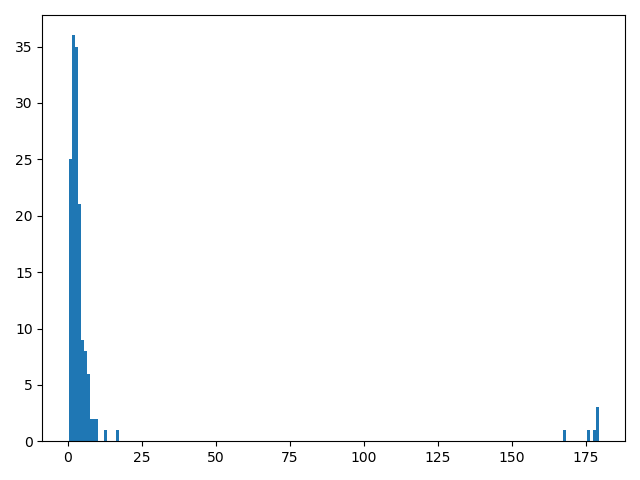} & 
    \includegraphics[width=0.30\linewidth]{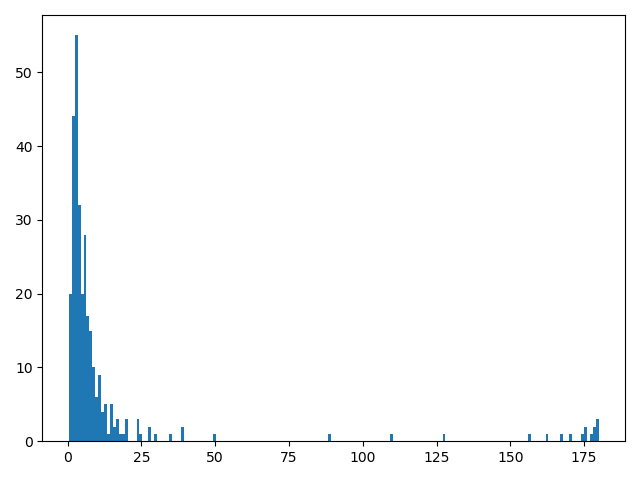} 
    \vspace{6mm} \\
    
    \small chair & \small diningtable & \small motorbike \\
    \includegraphics[width=0.30\linewidth]{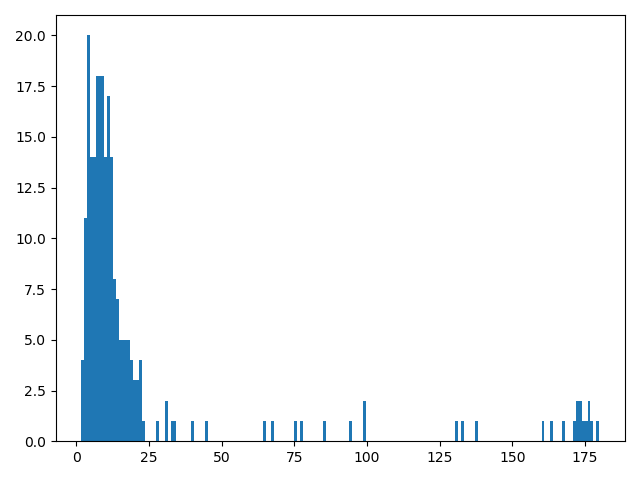} & 
    \includegraphics[width=0.30\linewidth]{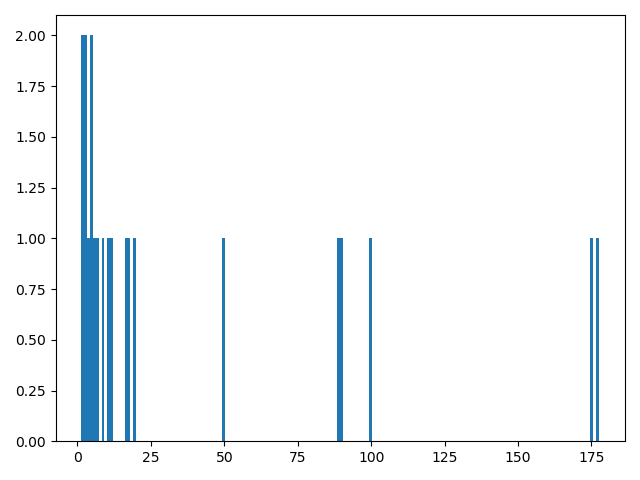} & 
    \includegraphics[width=0.30\linewidth]{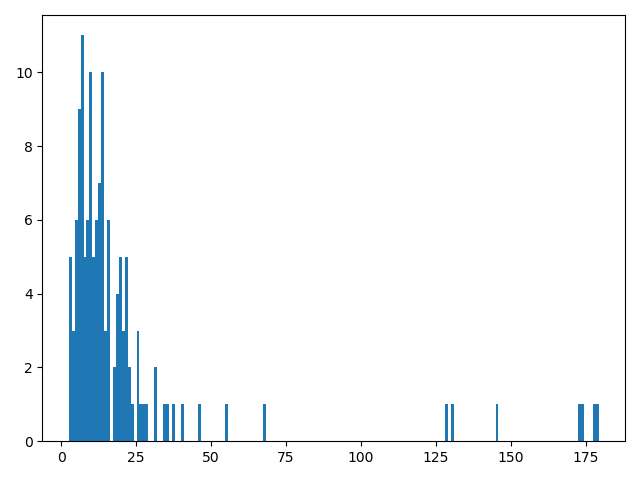} 
    \vspace{6mm} \\
    
    \small sofa & \small train & \small tvmonitor \\
    \includegraphics[width=0.30\linewidth]{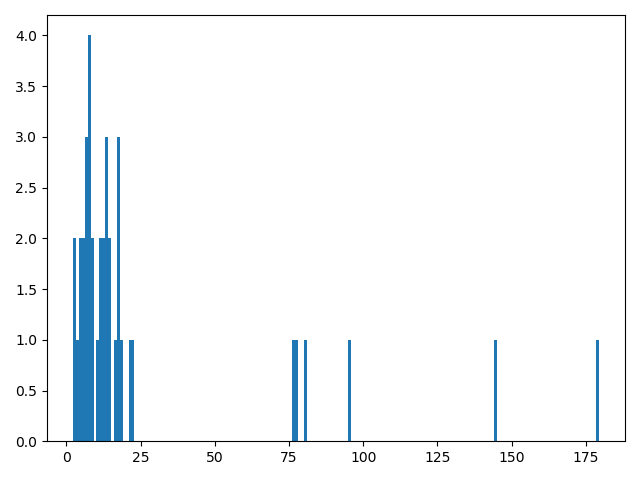} & 
    \includegraphics[width=0.30\linewidth]{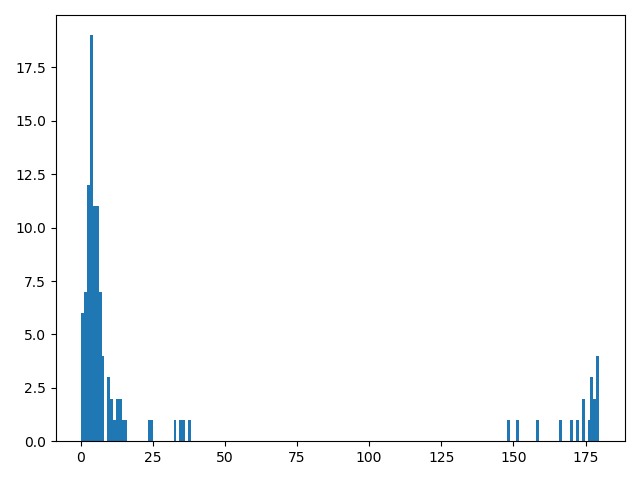} & 
    \includegraphics[width=0.30\linewidth]{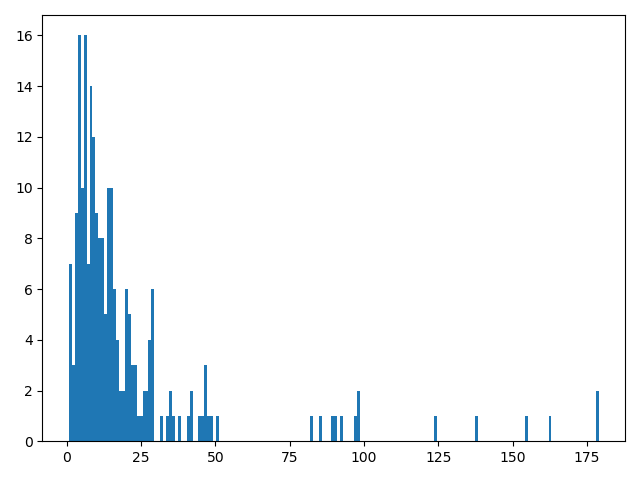} \\
    
    \end{tabular}
    \caption{\bf Histograms of azimuth angle prediction errors on the 12 object classes of Pascal3D+~\cite{xiang2014pascal3d}.}
    \label{fig:hist_azi}
\end{figure*}

Finally, we present in Figure~\ref{fig:hist_azi} the histograms of azimuth angle prediction errors on Pascal3D+~\cite{xiang2014pascal3d}. As shown in the figure, the largest viewpoint prediction errors come from the ambiguity caused by the symmetric objects, e.g., two-fold symmetric for `boat' and four-fold symmetric for `diningtable'.

\end{document}